\documentclass[a4paper,fleqn]{cas-sc}

\usepackage[authoryear]{natbib}
\usepackage{amsmath}
\usepackage{graphicx}
\usepackage{subcaption}

\usepackage[pagewise]{lineno}

\allowdisplaybreaks[1]
\def\tsc#1{\csdef{#1}{\textsc{\lowercase{#1}}\xspace}}
\tsc{WGM}
\tsc{QE}
\tsc{EP}
\tsc{PMS}
\tsc{BEC}
\tsc{DE}

\begin{document}
\let\WriteBookmarks\relax
\def\floatpagepagefraction{1}
\def\textpagefraction{.001}
\shorttitle{TrajGAIL: Generating Urban Vehicle Trajectories using Generative Adversarial Imitation Learning}
\shortauthors{S. Choi et~al.}

\title [mode = title]{TrajGAIL: Generating Urban Vehicle Trajectories using Generative Adversarial Imitation Learning}


\fntext[1]{This research was supported by the Ministry of Land, Infrastructure, and Transport (MOLIT, KOREA) Project ID: 21TLRP-B146733-04, Project Name: Connected and Automated Public Transport Innovation (National R\&D Project).}
\fntext[2]{All implementations of the model and the training algorithm is available at \url{https://github.com/benchoi93/TrajGAIL}.}
\fntext[3]{The authors appreciate Seoul Metropolitan Government and Dr. Min Ju Park for providing the taxi DTG data of Seoul city.}

\author[1]{Seongjin Choi}[orcid=0000-0001-7140-537X]
\ead{benchoi93@kaist.ac.kr}
\credit{Conceptualization of this study, Methodology, Software, Validation, Formal analysis,  Writing - original draft}

\author[2]{Jiwon Kim}[orcid=0000-0001-6380-3001]
\ead{jiwon.kim@uq.edu.au}
\credit{Conceptualization of this study, Methodology, Formal analysis, Supervision, Writing - review and editing.}

\author[1]{Hwasoo Yeo}[orcid=0000-0002-2684-0978]
\cormark[1]
\ead{hwasoo@kaist.ac.kr}
\credit{Data curation, Resources, Supervision, Funding acquisition, Project administration, Writing - review and editing}

\address[1]{Department of Civil and Environmental Engineering, Korea Advanced Institute of Science and Technology, 291 Daehak-ro, Yuseong-gu, Daejeon, Republic of Korea}
\address[2]{School of Civil Engineering, The University of Queensland, Brisbane St Lucia, Queensland, Australia}

\cortext[cor1]{Corresponding author}










\begin{abstract}
Recently, an abundant amount of urban vehicle trajectory data has been collected in road networks. 
Many studies have used machine learning algorithms to analyze patterns in vehicle trajectories to predict location sequences of individual travelers. 
Unlike the previous studies that used a discriminative modeling approach, this research suggests a generative modeling approach to learn the underlying distributions of urban vehicle trajectory data. 
A generative model for urban vehicle trajectories can better generalize from training data by learning the underlying distribution of the training data and, thus, produce synthetic vehicle trajectories similar to real vehicle trajectories with limited observations. 
Synthetic trajectories can provide solutions to data sparsity or data privacy issues in using location data. 
This research proposes \textit{TrajGAIL}, a generative adversarial imitation learning framework for the urban vehicle trajectory generation. 
In TrajGAIL, learning location sequences in observed trajectories is formulated as an imitation learning problem in a partially observable Markov decision process. 
The model is trained by the generative adversarial framework, which uses the reward function from the adversarial discriminator. 
The model is tested with both simulation and real-world datasets, and the results show that the proposed model obtained significant performance gains compared to existing models in sequence modeling.
%
%
\end{abstract}

\begin{graphicalabstract}
\includegraphics[width=\textwidth]{img/ModelFrame.PNG}
\end{graphicalabstract}

\begin{highlights}
\item Modeling urban vehicle trajectory generation as a partially observable Markov decision process.
\item A generative adversarial imitation learning framework for urban vehicle trajectory generation
\item Performance evaluation to assess both trajectory-level similarity and distributional similarity of datasets.
\end{highlights}

\begin{keywords}
Urban vehicle trajectories\sep 
Trajectory data analysis\sep 
Trajectory data generation\sep
Generative model\sep 
Generative adversarial imitation learning 
\end{keywords}


\maketitle
\section{Introduction}
Rapid advancements in location sensing and wireless communication technology enabled us to collect and store a massive amount of spatial trajectory data, which contains geographical locations of moving objects with their corresponding passage times \citep{lee2011trajectory}. Over the last decade, considerable progress is made in collecting, pre-processing, and analyzing trajectory data. Also, the trajectory data analysis is applied in various research areas, including behavioral ecology \citep{de2019trajectory}, transportation engineering \citep{wu2018location}, and urban planning \citep{laube2014computational}.
    
In transportation engineering, urban vehicle trajectory data are collected based on the location sensors installed inside vehicles or at the roadside and analyzed with various methods. The high-resolution mobility data of individual users in urban road networks offer unprecedented opportunities to understand vehicle movement patterns in urban traffic networks. It provides rich information on both aggregated flows and disaggregated travel behaviors. The aggregated flows include the origin-destination (OD) matrix and cross-sectional link traffic volumes. The disaggregated travel behaviors include user-centric travel experiences, namely, speed profile, link-to-link route choice behavior and travel time experienced by individual vehicles, as well as system-wide spatio-temporal mobility patterns, such as origin-destination pairs, routing pattern distributions, and network traffic states \citep{kim2015spatial}.
    

Most of the studies in the vehicle trajectory data analysis use machine learning methods. The recurrent neural network, for example, is used by many previous researchers due to its ability to learn sequential information in trajectory data. In machine learning, there are mainly two approaches to modeling: the \textit{discriminative} and the \textit{generative} modeling. A discriminative model learns a direct map from input $X$ to output (label) $Y$ or posterior probability $P(Y|X)$, which is the conditional probability of each label $Y$ given the input variable $X$. It only learns the decision boundaries between labels and does not care about the underlying distribution of data. In contrast, a generative model captures the underlying probability distribution, i.e., joint probability $P(X, Y)$, from which $P(Y|X)$ can be computed. One advantage of a generative model is that we can generate new (synthetic) data similar to existing data by sampling from $P(X, Y)$. 

Synthetic data generation based on generative models has gained increasing importance as the data generation process plays a significant role in various research fields in an era of data-driven world \citep{popic2019data}. It is mainly used to serve two purposes. The first purpose is to deal with the lack of real data. In many research fields, data collection is costly, and, therefore, it is often difficult to collect enough data to properly train and validate models. In this case, it is useful to generate synthetic data that are similar to the real observations to increase training and test samples. The second purpose is to address the issue with the privacy and confidentiality of real data. Many types of data contain personal information, such as gender, name, and credit card usage. Synthetic data can be combined with or replace such privacy-sensitive data with a reasonable level of similarity, thereby protecting privacy while serving the intended analysis.

Urban vehicle trajectory analysis has both challenges: data sparsity and data privacy issues. Although the sources and availability of urban trajectory data are increasing, most of the currently available trajectory datasets cover only a portion of all vehicles in the network. From network management and operations perspectives, there is a desire to infer vehicle trajectories that represent the whole population to have a more complete view of traffic dynamics and network performance. Moreover, urban vehicle trajectory data may contain personal information of individual drivers, which poses serious privacy concerns in relation to the disclosure of private information to the public or a third party \citep{chow2011privacy}. The ability to generate synthetic trajectory data that can realistically reproduce the population mobility patterns is, therefore, highly desirable and expected to be increasingly beneficial to various applications in urban mobility.


While synthetic trajectory data generation is a relatively new topic in transportation research communities, there are several existing research areas that have addressed similar problems. 
One example is \textit{trajectory reconstruction}. When two points in a road network are given as an initial point (treated as sub-origin) and a target point (treated as sub-destination), the models reconstruct the most plausible route between the two points. The trajectory reconstruction can be considered as generating trajectories between sub-origins and sub-destinations. Previous studies such as \cite{chen2011discovering} and \cite{hu2018graph} investigated on discovering the most popular routes between two locations. \cite{chen2011discovering} first constructs a directed graph to simplify the distribution of trajectory points and used the Markov chain to calculate the transfer probability to each node in the directed graph. The transfer probability is used as an indicator to reflect how popular the node is as a destination. The route popularity is calculated from the transfer probability of each node. \cite{hu2018graph} also used a graph-based approach to constructing popular routes. The check-in records which contain the route's attributes are analyzed to divide the whole space into zones. Then, the historical probability is used to find the most plausible zone sequences. 
Also, \cite{feng2015vehicle} and \cite{rao2018origin} estimated origin-destination patterns by using trajectory reconstruction. Both studies used particle filtering to reconstruct the vehicle trajectory between two points in automatic vehicle identification data. The reconstructed vehicle trajectory is then used to estimate the real OD matrix of the road network. 
Another problem that is relevant to trajectory generation is the \textit{next location prediction} problem, where the "next location" of a subject vehicle is predicted based on the previously visited locations of the subject vehicle. 
\cite{monreale2009wherenext}, for example, presented a decision tree to predict the next location based on the previously visited locations. Decision-tree based models, however, occasionally overfit the training dataset and lack the generalization ability to produce diverse trajectory patterns. \cite{gambs_next_2012} used Mobility Markov chain (MMC) to predict the next location among the clustered points or Point-of-Interests (POIs). 
The POIs considered in \cite{gambs_next_2012} are home, work, and other activity locations to model human activity trajectories throughout the day, rather than vehicle movement trajectories reflecting link-to-link vehicle driving behavior considered in this study. 
\cite{choi2019real} used a feed-forward neural network to predict the next intersection in a grid-structured road network. A set of intersections in Brisbane, Australia are treated as POI's to capture the link-to-link route choice behavior. \cite{jin2019augmented} used an augmented-intention recurrent neural network model to predict locations of vehicle trajectories of individual users. \cite{jin2019augmented} incorporated additional information on individual users' historical records of frequently visited locations into a next location prediction model. The past visited locations in historical records are represented as edge-weighted graph, and graph convolution network is used to incorporate this information into trajectory prediction. In \cite{choi2018network}, an urban road network is partitioned into zones based on the clustering of trajectory data points, and the prediction model based on recurrent neural network (RNN) is proposed to predict the zone that the subject vehicle would visit. \cite{choi2019attention} extended the idea of predicting the next zone and used network traffic state information to improve the RNN model's performance.






In fact, the existing models developed for the next location prediction problem can be applied for synthetic trajectory data generation. By sequentially applying the next location predictions, a synthetic vehicle trajectory can be generated. However, most of the existing models for next location prediction adopt a discriminative modeling approach, where the next locations are treated as labels and the model is trained to predict one or two next locations. The discriminative models have limitations in generating full trajectories, especially when sample trajectory data are sparse. it is only the decision boundaries between the labels that the models are trained to predict, not the underlying distributions of data that allow proper generalization for sampling realistic trajectories. As a result, it is necessary to develop a model based on the generative modeling approach to successfully perform synthetic trajectory data generation.


%
%


In this paper, we apply \textit{imitation learning} to develop a generative model for urban vehicle trajectory data. Imitation learning is a sub-domain of reinforcement learning for learning sequential decision-making behaviors or "policies". Unlike reinforcement learning that uses "rewards" as signals for positive and negative behavior, imitation learning directly learns from sample data, so-called "expert demonstrations," by imitating and generalizing the expert' decision-making strategy observed in the demonstrations. If we consider an urban vehicle trajectory as a sequence of decisions for choosing road links along a travel path, imitation learning can be applied to develop a generator that can reproduce synthetic data by imitating the decision-making process (i.e., driver' route choice behavior) demonstrated in the observed trajectory dataset. One approach to imitation learning is called \textit{Inverse Reinforcement Learning} (IRL), which aims to recover a reward function that explains the behavior of an expert from a set of demonstrations. Using the recovered expert reward function as feedback signals, the model can generate samples similar to the expert' decisions through reinforcement learning. \cite{ziebart2008maximum} and \cite{ziebart2008navigate} used maximum entropy IRL (MaxEnt) to generate synthetic trajectories similar to a given taxi dataset. One of the advantages of using IRL is that the model generates trajectories using both current states and expected returns of future states to determine an action---as opposed to considering only the knowledge up to the current state (e.g., previous visited locations)---, thereby enabling a better generalization of travel behavior along the whole trajectory.

Recently, there have been remarkable breakthroughs in generative models based on deep learning. In particular, \cite{goodfellow2014generative} introduced a new generative model called Generative Adversarial Networks (GAN), which addressed inherent difficulties of deep generative models associated with intractable probabilistic computations in training. GANs use an adversarial \textit{discriminator} to distinguish whether a sample is from real data or from synthetic data generated by the \textit{generator}. The competition between the generator and the discriminator is formulated as a minimax game. As a result, when the model is converged, the optimal generator would produce synthetic sample data similar to the original data. The generative adversarial learning framework is used in many research fields such as image generation \citep{radford2015unsupervised2}, audio generation \citep{oord2016wavenet}, and molecular graph generation \citep{de2018molgan}. 

GANs have been also applied in transportation engineering. \cite{zhang2019novel} proposed trip travel time estimation framework called \textit{T-InfoGAN} based on generative adversarial networks. They used a dynamic clustering algorithm with Wasserstein distance to make clusters of link pairs with similar travel time distribution, and they applied Information Maximizing GAN (InfoGAN) to travel time estimation. \cite{xu2020ge} proposed Graph-Embedding GAN (GE-GAN) for road traffic state estimation. Graph embedding is applied to select the most relevant links for estimating a target link and GAN is used to generate the road traffic state data of the target link. In \cite{li2020coupled}, GAN is used as a synthetic data generator for GPS data and travel mode label data. To solve the sample size problem and the label imbalance problem of a real dataset, the authors used GAN to generate fake GPS data samples of each travel mode label to obtain a large balanced training dataset.
The generative adversarial learning framework is also used for synthetic trajectory generation. \cite{liu2018trajgans} proposed a framework called trajGANs. Although this paper does not include specific model implementations, it discusses the potential of generative adversarial learning in synthetic trajectory generation. Inspired by \cite{liu2018trajgans}, \cite{rao2020lstm} proposed LSTM-TrajGAN with specific model implementations. The generator of LSTM-TrajGAN is similar to RNN models adopted in the next location prediction studies. 


This study proposes \textit{TrajGAIL}, a generative adversarial imitation learning (GAIL) model for urban vehicle trajectory data. GAIL, proposed by \cite{ho2016generative}, uses a combination of IRL's idea that learns the experts' underlying reward function and the idea of the generative adversarial framework. GAIL effectively addresses a major drawback of IRL, which is high computational cost. However, the standard GAIL has limitations when applied to the vehicle trajectory generation problem because it is based on the IRL concept that only considers a vehicle's current position as states in modeling its next locations \citep{ziebart2008maximum,ziebart2008navigate,zhang2019unveiling}, which is not realistic as a vehicle's location choice depends on not only the current position but also the previous positions. To overcome these limitations, this study proposes a new approach that combines a partially-observable Markov decision process (POMDP) within the GAIL framework. POMDP can map the sequence of location observations into a latent state, thereby allowing more generalization of the state definition and incorporating the information of previously visited locations in modeling the vehicle's next locations. In summary, the generation procedure of urban vehicle trajectories in TrajGAIL is formulated as an imitation learning problem based on POMDP, which can effectively deal with sequential data, and this imitation learning problem is solved using GAIL, which enables trajectory generation that can scale to large road network environments.



This paper is organized as follows. Section \ref{sec:method} describes the methodology of this paper. A detailed problem formulation is presented in Section \ref{ProbForm}, and the proposed framework of TrajGAIL is presented in Section \ref{sec:framework}. Section \ref{sec:eval} describes how the performance of the proposed model is evaluated. Section \ref{sec:data} introduces the data used in this study, and Section \ref{sec:baseline} introduces the baseline models for performance comparison. In Section \ref{sec:result}, the evaluation results are presented at both trajectory-level and dataset-level. Finally, Section \ref{sec:conclusion} presents the conclusions and possible future research.

\section{Methodology}\label{sec:method}

The objective of TrajGAIL is to generate location sequences in urban vehicle trajectories that are similar to real vehicle travel paths observed in a road traffic network. Here, the "similarity" between the real vehicle trajectories and the generated vehicle trajectories can be defined from two different perspectives.  First, the \textit{trajectory-level} similarity measures the similarity of an individual trajectory to a set of reference trajectories. For instance, the probability of accurately predicting the next locations---single or multiple consecutive locations as well as the alignment of the locations---are examples of trajectory-level similarity measures. Second, the \textit{dataset-level} similarity measures the statistical or distributional similarity over a trajectory dataset. This type of measure aims to capture how closely the generated trajectory dataset matches the statistical characteristics such as OD and route distributions in the real vehicle trajectory dataset. In this section, we present the modeling framework of TrajGAIL, where the procedure of driving in a road network is formulated as a partially observable Markov decision process to generate realistic synthetic trajectories, taking into account the similarities defined above.

%

\subsection{Problem Formulation}\label{ProbForm}






Let $Traj = \left\{(x_1,y_1,t_1), \cdots ,(x_N,y_N,t_N)\right\}$ be an urban vehicle trajectory, where $(x_i,y_i,t_i)$ is the $(x,y)$-coordinates and timestamp $t$ for the $i^{th}$ point of the trajectory, and $LocSeq = \left\{(x_1,y_1), \cdots ,(x_N,y_N)\right\}$ be the location sequence of $Traj$. When location points $(x,y)$ are continuous latitude and longitude coordinates, it is necessary to pre-process these coordinates and match them to a predefined set of discrete locations. Previous studies used different ways of defining discrete locations. For instance, \cite{choi2018network}, \cite{choi2019attention}, and \cite{ouyang2018non} used partitioned networks, so-called cells or zones, while \cite{choi2019real} and \cite{ziebart2008maximum} used road links to represent trajectories. In this paper, we represent a trajectory as a sequence of links to model link-to-link route choice behaviors in urban road networks. The location sequence of each vehicle trajectory is, thus, transformed to a sequence of link IDs by link matching function $f_M$: 

\begin{align}
    & LinkSeq = \big(l_1,\cdots,l_M \big) = f_M\big(LocSeq= \left\{(x_1,y_1), \cdots ,(x_N,y_N)\right\}\big) &
\end{align}

where $l_j$ is the link ID of the $j^{th}$ visited link along the trajectory. The goal of this study is to generate the link sequence of a trajectory by modeling and learning the probability distribution of $LinkSeq$, $P\big(LinkSeq\big) = P\big(L_1=l_1,\cdots,L_M=l_M \big)$ for a discrete random variable $L$ in all possible set of link IDs. Modeling this joint probability distribution is, however, extremely challenging, as also noted in the previous studies \citep{choi2018network,ouyang2018non}. A way to resolve this problem is to use a sequential model based on the Markov property, which decomposes the joint probability to the product of conditional probabilities as follows:


\begin{equation}
\begin{split}
P\big(LinkSeq\big) 
& = P\big(L_1 = l_1,\cdots,L_M=l_M \big)\\
& = P\big(L_M=l_M | L_{M-1} = l_{M-1} , \cdots , L_1=l_{1}\big) \times \cdots \times P\big(L_2=l_2| L_1=l_{1}\big) \cdot P\big( L_1=l_{1}\big) \\
& = P\big(L_M=l_M | L_{M-1}=l_{M-1}\big) \times \cdots \times P\big(L_2=l_2 | L_1=l_{1}\big) \times P\big( L_1=l_{1}\big) 
\end{split}
\end{equation}


The problem of modeling vehicle trajectories using this Markov property can be formulated as a Markov Decision Process (MDP). An MDP is a discrete-time stochastic control process based on the Markov property \citep{howard1960dynamic}. This process provides a mathematical framework for modeling sequential decision making of an agent. An MDP is defined with four variables: $(S,A,T,R)$, where $S$ is a set of states that the agent encounters, $A$ is a set of possible actions, $T(s,a,s')$ is a transition model determining the next state ($s' \in S$) given the current state ($s \in S$) and action ($a \in A$), and $R(s,a)$ is a reward function that gives the agent the reward value (feedback signal) of its action given the current state. If the transition is stochastic, transition model $T(s,a,s')$ can also be denoted as $P(s'|s,a)$. A policy $(\pi_{\theta})$ is defined as a $\theta$-parameterized function that maps states to an action in the deterministic case $(\pi_{\theta} (s) \rightarrow a)$, or a function that calculates the probability distribution over actions $\big(\pi_{\theta} (s) = P(a|s)\big)$ in the stochastic case. The objective of MDP's optimization is to find the optimal policy that maximizes the expected cumulative rewards, which is expressed as:


\begin{align}
\label{eq:maxreward}
    & \pi_{\theta^*} = \arg \max_\theta \mathbf{E}\Big[\sum_{t=0}^{\infty} \gamma^{t} \cdot R(s,a)\Big] & 
    & a \sim Categorical \Big(\pi_{\theta} (s)=P(a|s) \Big)&
\end{align}
where $ \pi_{\theta^*}$ is the optimal policy with parameter $\theta^*$, $a$ is a sampled action from $\pi_{\theta} (s)= P(a|s)$ (in this study, we use discrete action space, so actions are sampled from categorical distribution), and $\gamma$ is the discount rate of future rewards.


How to define the four variables of MDP is critical to the successful training of a policy model. The states $(s)$ should incorporate enough information so that the next action $(a)$ is determined based only on the current state $(s)$, and the transition model $T$ should correctly reflect the transition of states in the environment it models. Finally, the reward function $R$ should give a proper training signal to the agent to learn the optimal policy.

In TrajGAIL, the vehicle movement in a road network is formulated as an MDP. We set road segments or links as states and transitions between links as actions. In this case, the transition model can be defined as a deterministic mapping function that gives the next link $(s')$ given the current link $(s)$ and the link-to-link movement choice $(a)$, i.e.,  $T:(s, a) \mapsto s'$. The policy represents a driver's route choice behavior associated with selecting the next link at each intersection. This road network MDP, thus, produces vehicle trajectories---more specifically, link sequences---as a result of sequential decision making modeled by this policy.

As mentioned above, MDP assumes that the action $(a)$ is determined based only on the current state $(s)$. However, it is likely that vehicles' link-to-link movement choice is affected by not only the current location but also the previous locations. Moreover, vehicle movements in a road network is a result of complex interactions between a large number of drivers and road environment such as the generation and distribution of trips and the assignment of the routes and, therefore, the link choice action cannot be determined solely by road segment information alone as a state. The model needs more information such as origin, destination, trip purpose, and the prevailing traffic state. Incorporating all such information in the state definition, however, makes the problem intractable due to an extremely large state space. It is, thus, desirable to relax the assumption such that the action is determined based on the current state as well as some \textit{unobservable} states.


\begin{figure}
  \centering
  \includegraphics[width=0.7\textwidth]{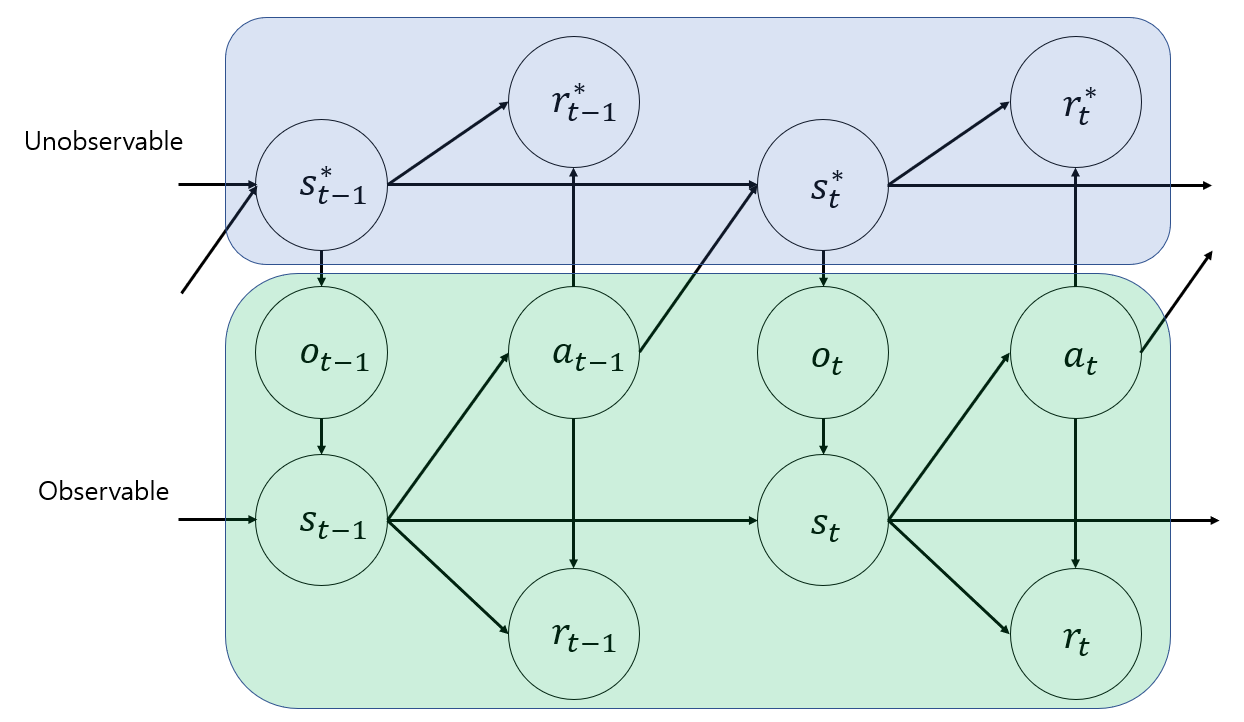}
  \caption{ Partially observable Markov Decision Process }
  \label{fig:pomdp}
\end{figure}

This can be achieved by employing a partially observable MDP (POMDP). A POMDP assumes that an MDP determines the model dynamics, but the agent cannot directly observe the underlying states. Instead of directly using the states as MDP, POMDP uses a surrogate state such as probability distribution over the set of possible states \citep{kaelbling1998planning} and belief state \citep{rao2010decision}. Figure \ref{fig:pomdp} shows the graphical model of POMDP. It is assumed that there exist latent unobservable states $s^* \in S^*$. We can only partially observe $s^*$ through observation $o \in O$. 
Using $o$ or sequence of $o$'s, latent state $s^*$ is estimated and this estimated latent state, or the belief state, is represented as $s \in S$.


As a result, instead of four variables of MDP, POMDP uses five variables $(O, S, A, T, R)$, where $O$ represents the set of possible observations. The belief state at time state $t$, $s_t$, is estimated based on the sequence of $o$ representing all observations up to the current time $ t $, which is assumed to be the estimate of the latent unobservable state, $ \hat{s}^*_t $, as follows: 


\begin{align}\label{eq:obtos}
  & s_t=f(o_1,...,o_t)=\hat{s}^*_t \qquad \qquad  s \in S, o \in O
\end{align}

In TrajGAIL, the observation space, $O$, is defined as the ID of links in the road network and two virtual tokens representing the start and the end of a trip $(Start, End)$. Actions $A$ are transitions between links. In \cite{ziebart2008maximum}, the set of actions includes all possible link transitions.
However, this can lead to a very large action space even with a moderate-sizes network with hundreds of links, requiring high computational cost. To reduce the computational complexity, we instead define a set of common actions that represent possible movements between two connected links, namely, $[Straight, Left, Right, Terminate]$, where $Straight$, $Left$, and $Right$ represent the movement direction at the end of each link (at intersections) and $Terminate$ represents the termination of a trip (i.e., a vehicle reached its destination). These four actions are sufficient for our current study as we consider a grid-structured network, where all intersections are four-way intersections. However, it is also possible to model general networks with more diverse intersection structures such as five-way or T-shape intersections as we can apply a "mask" that allows flexibility to further define specific actions available for each link, which would be a subset of the network-wide common action set. For instance, one can define six actions for a network with the maximum intersection size of six and specify only a subset of available actions for each link if it has less than six connected roads.

To summarize, we formulate a partially observable Markov Decision Process to develop a generative model for vehicle trajectories, which produces the optimal policy describing optimal actions given a sequence of observations. 

\subsection{Model Framework}\label{sec:framework}

\subsubsection{Preliminaries and Background - Imitation Learning}
In this study, the imitation learning framework is used to develop a generative model represented in POMDP formulation. Imitation learning is a learning problem that aims to train a model that can act like a given expert. Usually, demonstrations of decisions of the expert are given as a training dataset. In this study, a real vehicle trajectory dataset serves as expert demonstrations so that the model learns the decision-making process of vehicle movements in a road network observed in the given dataset.
%
%
There are mainly two categories of approaches in imitation learning: \textit{behavior cloning} and \textit{inverse reinforcement learning}. 

Behavior cloning considers the imitation learning problem as a supervised learning problem. In behavior cloning, given the expert demonstrations, the state and action sequence is divided into independent state-action pairs and a model is trained to directly learn the relationship between input (state) and output (action) based on these sample pairs. The biggest advantage of behavior cloning is simplicity. However, because of its simplicity, the model fails to make proper generalization in complex tasks. Simple generative models based on Markov Chain \cite{gambs2010show} and Recurrent Neural Networks \cite{choi2018network,choi2019attention,liu2016predicting} can be classified into this category of imitation learning.

The inverse reinforcement learning (IRL) uses an indirect approach.
The objective of IRL is to find the reward function that the agent is optimizing given the measurements of agents' behavior and sensory inputs to the agents \citep{russell1998learning}.
It is assumed that the experts follow certain rules known as a reward function. The main idea of IRL is to learn this reward function to imitate the experts based on the history of experts' behaviors in certain situations. It is called "inverse" reinforcement learning because it learns the reward function that represents the experts' decisions from their states and actions, whereas the reinforcement learning (RL) learns to generate states and actions from a given reward function. Some of the key papers on IRL problems include \cite{ng2000algorithms, abbeel2004apprenticeship, ziebart2008maximum, wulfmeier2015deep, ho2016generative}, which readers are referred to for more details on IRL.

Given expert policy $\pi_E$, the objective 
of IRL is 
to find a reward function ($r$) that maximizes the difference between the expected rewards from the expert and the RL agent ($\mathbf{E}_{\pi_E} [r(s,a)] - RL(r)$) such that the expert performs better than all other policies \citep{ho2016generative}. This is achieved by minimizing the expected reward from RL agent ($RL(r)$) and by maximizing the expected reward from the expert ($\mathbf{E}_{\pi_E} [r(s,a)]$), while minimizing the reward regularizer ($\psi(r)$). On the other hand, when a reward function ($r$) is given, the objective of RL is to find a policy ($\pi$) that maximizes the expected reward ($\mathbf{E}_\pi[r(s,a)]$) while maximizing the entropy of the policy ($H(\pi)$).

\begin{align}\label{eq:irl}
    & IRL(\pi_E) = \underset{r \in \mathbf{R}}{\arg\min} \Big( \psi(r) +  RL(r) - \mathbf{E}_{\pi_E} [r(s,a)]  \Big) &
\end{align}

\noindent
where,

\begin{align}
    & RL(r) =   \max_{\pi \in \Pi} \Big( H(\pi) + \mathbf{E}_\pi[r(s,a)] \Big) &
  \label{eq:rl}
\end{align}

\noindent
where 
$\mathbf{R}$ is the largest possible set of reward functions ($\mathbf{R} = \{ r:S \times A \longrightarrow \mathbb{R} \} $), 
$\psi(r)$ is the convex reward function regularizer, and $H(\pi) = \mathbf{E}_\pi [- \log \pi(a|s)]$  is the causal entropy of the policy $\pi$ \citep{ho2016generative}.

It is interesting to investigate the relationship between $RL$ and $IRL$ in Eq. (\ref{eq:irl}) and Eq. (\ref{eq:rl}). $RL$ tries to find the optimal policy $\pi$ that maximizes the expected rewards, and $IRL$ tries to find the optimal reward function that maximizes the difference between expert policy ($\pi_E$) and $RL$'s policy ($\pi$). In some sense, $RL$ can be interpreted as a generator that creates samples based on the given reward, and $IRL$ can be interpreted as a discriminator that distinguish the expert policy from $RL$'s policy. This relationship is similar to the framework of Generative Adversarial Networks (GAN). GANs use an adversarial \textit{discriminator} ($D$) that distinguishes whether a sample is from real data or from synthetic data generated by the \textit{generator} ($G$). The competition between generator and discriminator is formulated as a minimax game. As a result, when the model is converged, the optimal generator would produce synthetic sample data similar to the original data. Eq. (\ref{eq:gan}) shows the formulation of minimax game between $D$ and $G$ in GANs.


\begin{align}\label{eq:gan}
    & \min_G \max_D \Big( \mathbf{E}_{x \sim p_{data}(x)} \big[\log D(x)\big] + \mathbf{E}_{z \sim p_{z}(z)} \big[\log \left(1-D(G(z)) \right) \big] \Big) &
\end{align}

With a proper selection of the regularizer $\psi(r)$ in \textit{IRL} formulation in Eq. (\ref{eq:irl}), \cite{ho2016generative} proposed \textit{generative adversarial imitation learning} (GAIL). The formulation of minimax game between the discirminator ($D$) and the policy ($\pi$) is shown in Eq. (\ref{eq:gail})


\begin{align}
    & \min_D \max_\pi \Big( \mathbf{E}_{\pi} \big[\log D(s,a)\big] + \mathbf{E}_{\pi_E} \big[\log \left(1-D(s,a) \right) \big] - \lambda H(\pi) \Big) &
  \label{eq:gail}
\end{align}

Eq. (\ref{eq:gail}) can be solved by finding a saddle point $(\pi, D)$. To do so, it is necessary to introduce function approximations for $\pi$ and $D$ since both $\pi$ and $D$ are unknown functions and it is very difficult, if not impossible, to define a exact function form for them. Nowadays, deep neural networks are widely used for function approximation. By computing the gradients of the objective function with respect to the corresponding parameters of $\pi$ and $D$, it is possible to train both generator and discriminator through backpropagation. In the implementation, we usually take gradient steps for $\pi$ and $D$ alternatively until both networks converge.

While GAIL provides a powerful solution framework for synthetic data generation, the original GAIL model \cite{ho2016generative} could not be directly used for our problem of vehicle trajectory generation. From our experiments, we found that the standard GAIL tends to produce very long trajectories with many loops, indicating that vehicles are constantly circulating in the network. This is because the generator in GAIL tries to maximize the expected cumulative rewards and creating a longer trajectory can earn higher expected cumulative rewards as there is no penalty of making a trajectory longer.

There are several ways to address this issue. Possible approaches include giving a negative reward whenever a link is visited to penalize a long trajectory or using positional embedding (the number of visited links) as state. However, a better approach would be to let the model know the vehicle's visit history (a sequence of links visited so far) and learn that it is unrealistic to visit the same link over and over. Our proposed TrajGAIL framework achieves this and addresses the limitation of GAIL in trajectory generation by assuming POMDP and using RNN embedding layer.

\subsubsection{TrajGAIL: Generative Adversarial Imitation Learning Framework for Vehicle Trajectory Generation}\label{sec:trajgail}

\begin{figure}
  \centering
  \includegraphics[width=\textwidth]{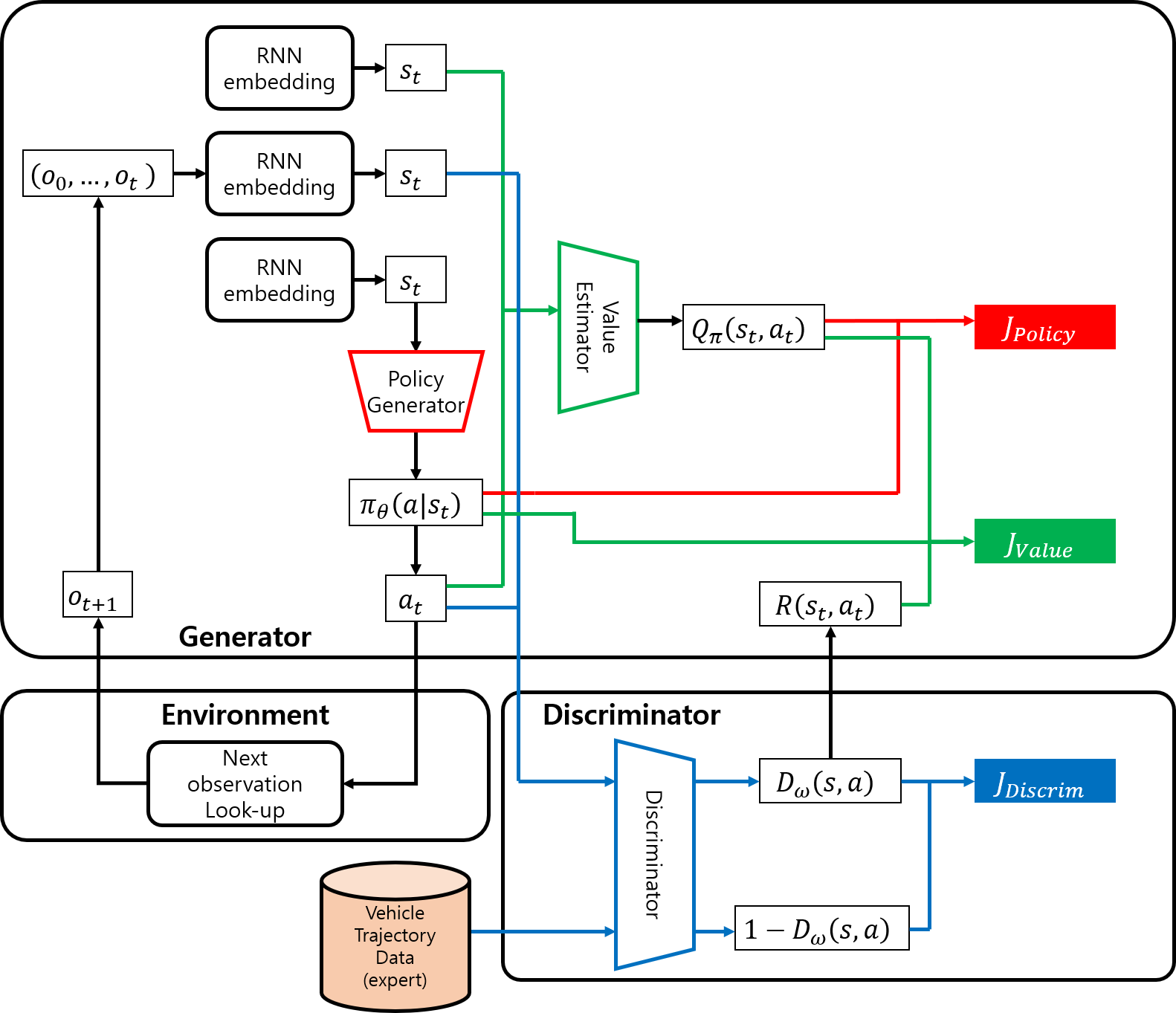}
  \caption{The model framework of TrajGAIL}
  \label{fig:gailframework}
\end{figure}

TrajGAIL uses POMDP to formulate vehicle trajectory generation as a sequential decision making problem and GAIL to perform imitation learning on this POMDP to learn patterns in observed trajectories to generate synthetic vehicle trajectories similar to real trajectories. 
In vehicle trajectory generation, it is important to take into account not only the previous locations of the trajectories, but also the expected future locations that the trajectory is expected to visit. By using POMDP, TrajGAIL considers how realistic the previously visited locations are. By using GAIL, TrajGAIL can also consider how realistic future locations would be because the imitation learning framework in GAIL uses an objective function to maximize the expected cumulative future rewards when generating new actions, which captures how realistic the remaining locations will be.
It is noted that, in this study, TrajGAIL focuses on generating location sequences (link sequences) of trajectories without considering time components. Throughout the paper, we use the term \textit{trajectory generation} to refer to the generation of link sequences representing trajectory paths for the sake of brevity. Figure \ref{fig:gailframework} shows the model framework of TrajGAIL. As in GAIL, TrajGAIL consists of the discriminator and the generator, where the discriminator gives reward feedback to the vehicle trajectories generated by the generator until both converge. The generator works as a reinforcement learning agent, and the discriminator works as an inverse reinforcement learning agent. Below we provide more details on each of these two modules.

\textbf{The Generator of TrajGAIL}. The primary role of the generator is to make realistic synthetic vehicle trajectories. The generator creates $N$ trajectories by a policy roll-out, or an execution of a policy from initial state ($Start$ of trip) to terminal state ($End$ of trip). A trajectory starts with the virtual token $Start$. By sequentially applying the policy generator until the current observation reaches the other virtual token $ End $, the generator produces a whole vehicle trajectory. As our problem is formulated as a POMDP, we need to map the sequence of observations into the latent states. \cite{rao2010decision} suggested that the belief states ($s_t$), the estimate of probability distribution over latent states, can be computed recursively over time from the previous belief state. The posterior probability of state $i$ at the $t$-th observation, denoted by $s_t(i)$ can be calculated as follows:

\begin{equation}
\begin{split}
    \label{eq:transition}
    s_t(i) & = P(s^*_t=i|o_t,a_{t-1},o_{t-1},\cdots,a_0,o_0) \\
          & \propto  P(o_t|s^*_t=i) \cdot P(s^*_t=i|a_{t-1},o_{t-1},\cdots,a_0,o_0) \\
          & \propto  P(o_t|s^*_t=i) \cdot \sum_j T(j,a_{t-1} , i) s_{t-1}(j) 
\end{split}
\end{equation}

\noindent
where $s^*_t$ is the latent states at the $t$-th observation including both observable and unobservable variables, $P(o|s^*)$ is the probability of observation $ o $ given $s^*$, and $T(s, a, s')$ is the transition model that maps the current state ($s$) and the action ($a$) to the next state ($s'$).

Eq. (\ref{eq:transition}) indicates that the current belief state vector $(s_t)$ is a combination of the information from the current observation $(P(o_t|s^*_t))$ and the feedback from the previous computation of the belief state $(s_{t-1})$. In \cite{rao2010decision}, the author recognized the similarity between the structure of this equation and recurrent neural networks (RNN) and suggested using RNN for belief state estimation. Many previous studies on the next location prediction problem suggest that RNNs show great performance in embedding the sequence of locations into a vector \citep{choi2018network, choi2019attention, feng2018deepmove}. Accordingly, we use an RNN embedding layer to map the sequence of observations (link IDs) to a belief state vector. Since the entire historical sequence is embedded in the current (belief) state via RNN embedding and the actions are still determined based only on the current state, the Markov assumption in MDP is not violated, while sequential information can be effectively captured within the model.

In the implementation of RNN embedding layer, the size of input tensor (observation sequence) is $[B\times L]$ and the size of output tensor (belief state vector) is $[B \times H]$, where $B$ is the batch size, $L$ is the maximum observation sequence length in the batch, and $H$ is the number of hidden neurons.

Based on the belief state vector $(s_t)$, the \textit{policy generator} within the TrajGAIL generator module calculates the probability of the next action $(\pi (a|s_t))$. The policy $\pi (a|s_t)$ has a size of $[B \times A]$, where $A$ is the size of action space. The next action is sampled from a multinomial distribution with the probability $(\pi (a|s_t))$. The next observation is determined by the \textit{next observation look-up} table in the road network environment, $T_o(o_t,a_t,o_{t+1})$, which maps the current observation ($o_t$) and the action ($a_t$) to the next observation ($o_{t+1}$). This process continues until the current observation reaches the virtual token $End$. 

In reinforcement learning, a \textit{value function} is often used to calculate the expected return of the actions at the current state. Here, we use a state-action value function $Q_{\pi} (s,a)$, which is estimated via the \textit{value estimator} in TrajGAIL's generator. The state-action value function, $Q_{\pi} (s,a)$, has size of $[B \times 1]$ since it represents a value scalar of each input observation sequence. The value estimator has a separate RNN embedding layer to process the sequence of observations into the belief state. Based on the processed belief state and a given action, the value estimator calculates the expected return of the action at current belief state. The estimated value, or the expected return, is used as a coefficient when updating the policy generator. If the estimated value of a given action is large, the policy generator model is reinforced to give the similar actions more often. This value estimator is also modeled as a deep neural network, which is trained to minimize the value objective function, $J_{Value}$, defined as follows:

\begin{equation}\label{eq:grad_value}
\begin{split}
J_{Value} & = \mathbf{E} \Big[ \big(   Q_{\pi} (s_t,a_t) - G(s_t,a_t)   \big)^2  \Big] & \\
            & = \mathbf{E} \Big[ \big(   Q_{\pi} (s_t,a_t) - (R(s_t,a_t) + \gamma \cdot \mathbf{E} [R(s_{t+1},a_{t+1})] +\gamma^2 \cdot \mathbf{E} [R(s_{t+2},a_{t+2})] + ...   \big)^2  \Big] \\
            & = \mathbf{E} \Big[ \big(   Q_{\pi} (s_t,a_t) - (R(s_t,a_t) + \gamma (\mathbf{E} [R(s_{t+1},a_{t+1})] +\gamma \cdot \mathbf{E} [R(s_{t+2},a_{t+2})] + ...)   \big)^2  \Big] \\
            & = \mathbf{E} \Big[ \big(   Q_{\pi} (s_t,a_t) - (R(s_t,a_t) + \gamma \cdot \mathbf{E}[Q_{\pi} (s_{t+1},a_{t+1})]   )   \big)^2  \Big] \\
            & = \mathbf{E} \Big[ \big(   Q_{\pi} (s_t,a_t) - (R(s_t,a_t) + \gamma \cdot \sum \pi(a_{t+1} | s_{t+1})\cdot Q_{\pi} (s_{t+1},a_{t+1})  )   \big)^2  \Big] 
\end{split}            
\end{equation}

where a mean squared error (MSE) loss between the value estimate $(Q_{\pi} (s,a))$ and the actual $\gamma$ discounted return $(G(s,a))$ is used. 

The objective of the policy update is to maximize the expected cumulative reward function as shown in Equation (\ref{eq:maxreward}). We define the policy objective as $J(\theta)$ and we maximize the policy objective to improve the policy generator at every iterations. 
In order to compute the gradient of the policy objective, 
we use the Policy Gradient Theorem \citep{sutton2000policy}.
In the Policy Gradient Theorem, 
for any differentiable $\theta$-parameterized policy $\pi_\theta$, the policy gradient of policy objective $\nabla_{\theta} J_{PG}(\theta)$ is given as:

\begin{align}
    & \nabla_{\theta} J_{PG}(\theta) = \mathbf{E} \Big[ \nabla_\theta \log \pi_{\theta} (a|s) \cdot Q_{\pi_\theta} (s,a) \Big] = \nabla_\theta J_{Policy} &
\end{align}

\noindent
where $J_{Policy} =  \mathbf{E} \Big[ \log \pi_{\theta} (a|s) \cdot Q_{\pi_\theta} (s,a) \Big]$.

Additionally, we add an entropy maximization objective \citep{ziebart2008maximum, ho2016generative} to the policy objective $J(\theta)$ in order to prevent the policy from converging to a local optimal policy. Often the global optimal policy is difficult to learn because of the sparsity of reward function. In this case, the policy would converge to a local optimal policy that only generates a limited variety of trajectories without considering the underlying distribution of actions. The entropy maximization objective helps counteract this tendency, guiding the policy to learn the underlying distribution of actions and eventually learn the underlying distribution of trajectories. As a result, we use the following equation to update the parameters of the policy generator to maximize the policy objective. 

\begin{align}\label{eq:grad_policy}
    %
\begin{split}
    & J(\theta) = J_{PG}(\theta) + \lambda H(\pi_\theta) \\
    & \nabla_{\theta} J(\theta) = \nabla_{\theta} J_{PG}(\theta) + \lambda \nabla_{\theta} H(\pi_\theta)= \nabla_{\theta} J_{Policy} + \lambda \nabla_{\theta} H(\pi_\theta)\\
\end{split}
\end{align}

\textbf{The Discriminator of TrajGAIL}. The discriminator solves the classification problem by distinguishing real vehicle trajectories from generated vehicle trajectories. 
As the generator gets improved to create more realistic vehicle trajectories, the discriminator's ability to classify the generated trajectories from the real trajectories is also improved through iterative parameter updates and fine-turning.
%
%
This competition of two neural networks is the fundamental concept of the generative adversarial learning framework. In the discriminator update step, the samples from the real vehicle trajectory dataset are labeled as 0, and the samples from the generator are labeled as 1. For both real and generated vehicle trajectories, we put the sequence observation and the action taken at the last observation as an input, process the sequence of observation into a belief state through an RNN embedding layer, and calculate the probability ($D_\omega (s,a)$) that the given sequence of observations and the action are from the generator. The discriminator probability, $D_\omega (s,a)$, has size of $[B \times 1]$ similar to $Q_{\pi} (s,a)$. The parameters of the discriminator are updated to minimize the binary cross-entropy loss. The $\omega$-parameterized discriminator is updated to minimize the discriminator objective, $J_{Discrim}$, with the following gradient term.

\begin{align}\label{eq:discrim}
    & \nabla_{\omega} J_{Discrim} =  \Bigg[\mathbf{E}_{ (s,a) \sim \pi_{\theta} } \Big[ \nabla_{\omega} \log(D_{\omega} (s,a)) \Big] + \mathbf{E}_{(s,a) \sim \pi_{E}} \Big[ \nabla_{\omega} \log(1- D_{\omega} (s,a)) \Big] \Bigg] &
\end{align}

The discriminator gives the training signal to the generator through the reward function ($R(s,a)$) as shown in Figure \ref{fig:gailframework}. The generator is trained to maximize the binary cross-entropy loss of the discriminator. As the second term of Eq. (\ref{eq:discrim}) is irrelevant to the parameters of the generator, the objective of the generator is to maximize the first term of Eq. (\ref{eq:discrim}). As a result, the reward function is defined as follows:

\begin{align}
    & R(s,a) = - \log (D_{\omega}(s,a)) &
\end{align}

\begin{figure}
  \centering
  \includegraphics[width=\textwidth]{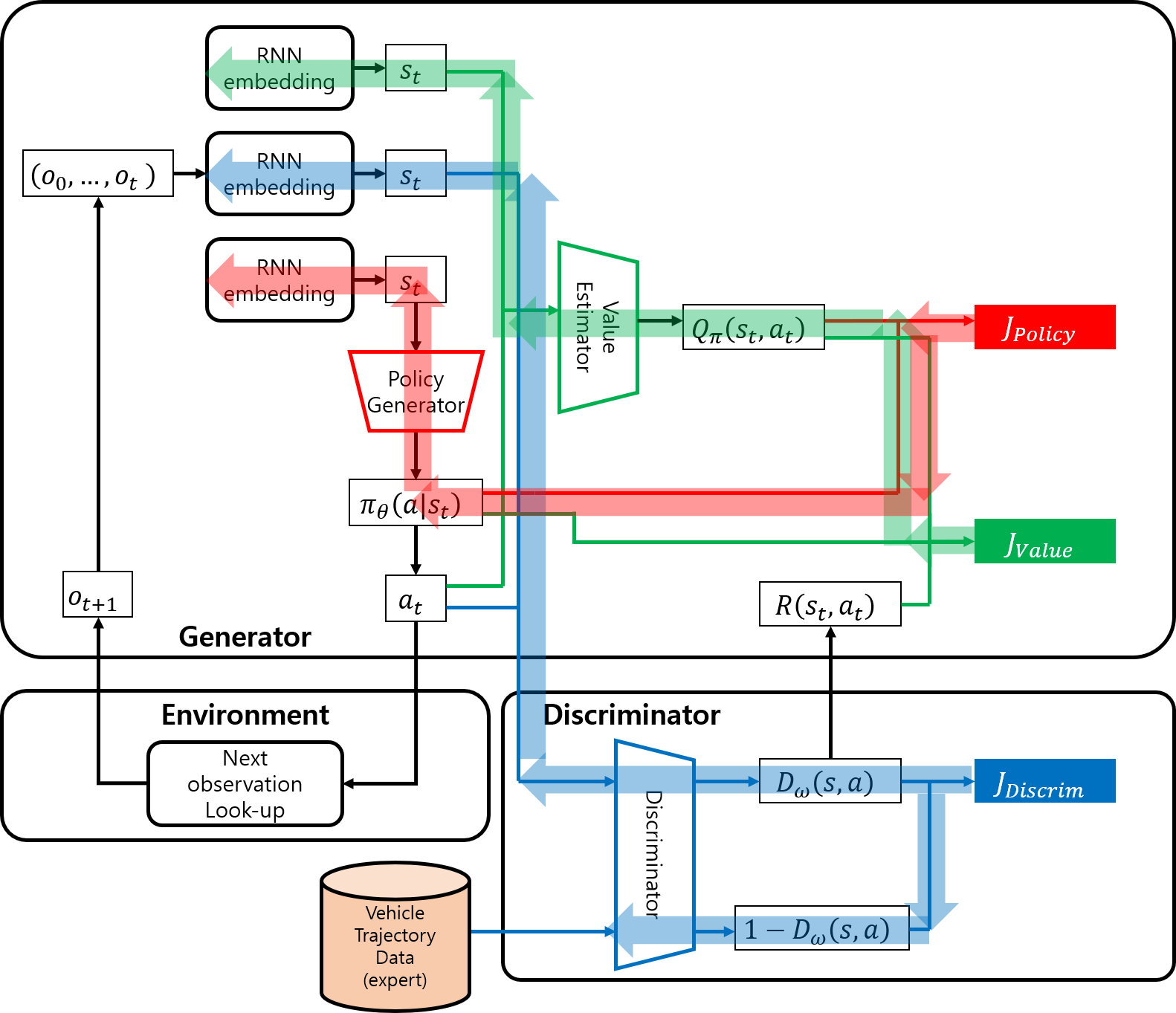}
  \caption{Backpropagation schema of TrajGAIL (red arrows for policy objective, green arrows for value objective, and blue arrows for discriminator objective)}
  \label{fig:gailbackprop}
\end{figure}

\textbf{Backpropagation}. Figure \ref{fig:gailbackprop} shows the schema of backpropagation to update parameters in the whole TrajGAIL framework. There are three different objective functions, $J_{Policy}$, $J_{Value}$, and $J_{Discrim}$, used to update parameters associated with policy generator, value estimator, and discriminator, respectively. Each objective is backpropagated to update only the related parameters in the deep neural networks. The backpropagation route of each objective function is indicated in different colors, i.e., red arrows for policy objective ($J_{Policy}$), green arrows for value objective ($J_{Value}$), and blue arrows for discriminator objective ($J_{Discrim}$). It is noted that the policy generator, the value estimator, and the discriminator use their own RNN embedding layers to embed the sequence of observations to the latent states. Using a separate RNN embedding layer for each of these three modules enables each module to interpret the sequence of observations and update RNN parameters in such a way as to maximize its performance in the task given to the module. For example, the discriminator might have a different interpretation on the sequence states from the policy generator, because the discriminator might focus more on the information on the whole sequence such as trajectory length to execute a discriminative task, while the policy generator might focus more on the current position to decide the next link-to-link transition. The backpropagation routes through the RNN embedding layers in Figure \ref{fig:gailbackprop} represent that each of these three modules updates its own RNN embedding layer. These RNN embedding layers play an integral role in combining the POMDP concept within the GAIL framework, which are the main distinction between TrajGAIL and stardard GAIL models.

\textbf{Training Techniques}. During the implementation of generative adversarial network algorithms, there are several techniques that can be used to avoid training failure and facilitate model convergence. The first technique used in this study is to maintain similar learning levels between the discriminator and the generator while training. If one overpowers the other, a proper competition cannot be formed and, thus, the model cannot learn from the competition mechanism of generative adversarial networks.
%
Options to balance the learning levels include setting different learning rates and/or using different numbers of update-steps for training the generator and the discriminator. In this study, we used the same learning rate for both discriminator and generator and used a different number of update-steps. At each iteration, the generator is updated six times, while the discriminator is updated twice, because the discriminator usually learned faster in this study. The second technique is to use a sufficient number of trajectories generated for training. At each iteration, a certain number of trajectories are created and used to update the parameters of the model. If the number of generated trajectories is small, the model may result in a problem called "mode collapse." The mode collapse is defined as a case where the generator collapses, producing a limited variety of data \citep{dumoulin2016adversarially,lin2018pacgan}. Sometimes the generator oscillates among a few data points without converging to the equilibrium. In our case, the mode collapse leads the generator to produce trajectories for a few specific routes only. Using a sufficient number of sample trajectories at each training iteration solved this problem in our case. The details of the sample size used in this study are provided in the description of the evaluation results below.

\section{Performance Evaluation}\label{sec:eval}

\subsection{Data}\label{sec:data}

The performance of TrajGAIL was evaluated based on two different datasets. The first dataset is a virtual vehicle trajectory dataset generated by a microscopic traffic simulation model, AIMSUN. The second dataset includes the data collected by the digital tachograph (DTG) installed in the taxis operating in Gangnam District in Seoul, South Korea.

\begin{figure}
  \centering
  \includegraphics[width=0.3\textwidth]{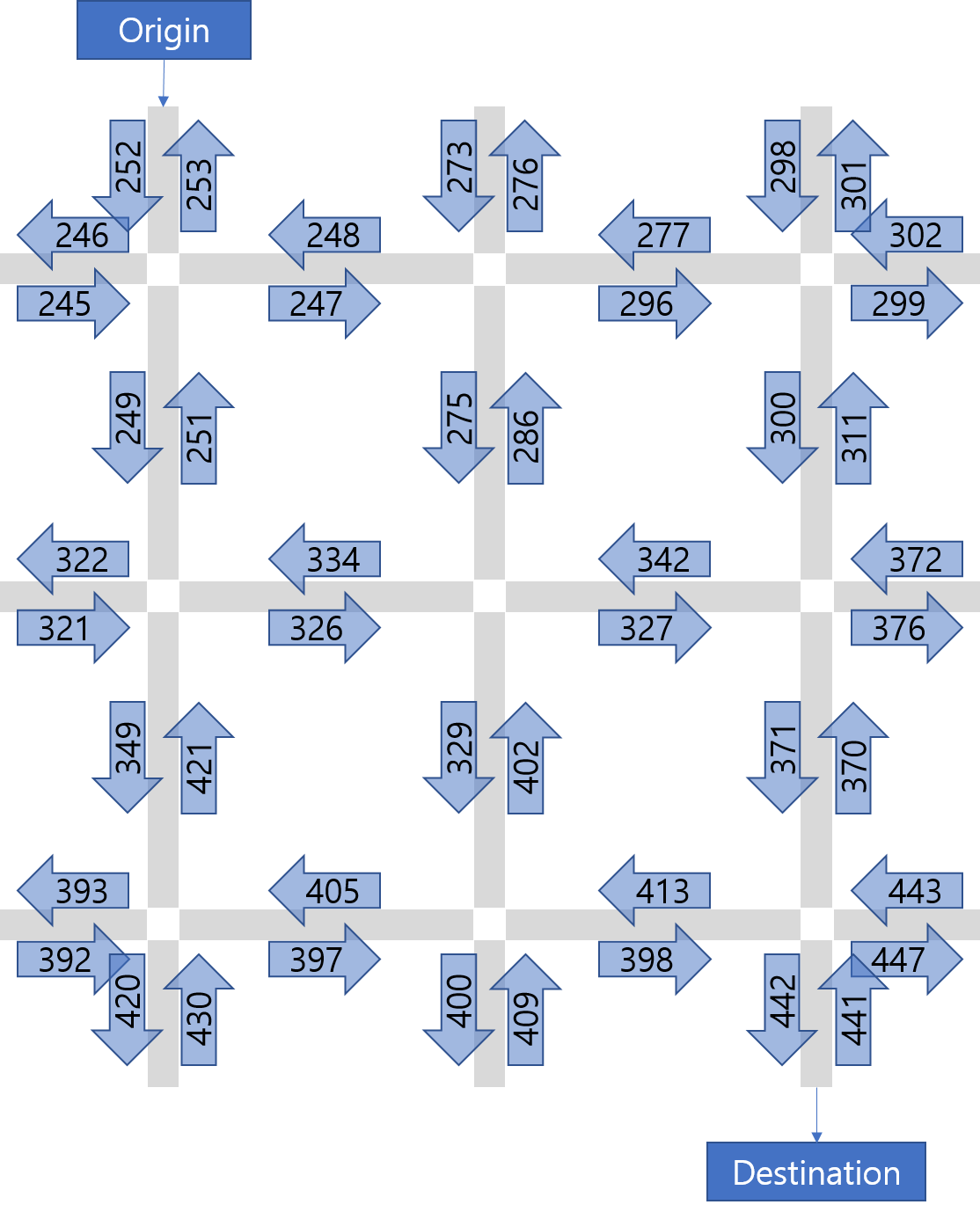}
      \caption{Simulated traffic network in AIMSUN environment. The locations of origin and destination is shown in the figure.}
  \label{fig:aimsun_single}
\end{figure}

AIMSUN uses dynamic traffic assignment \citep{barcelo2005dynamic} to select the appropriate route for each vehicle. We can select five different route choice models: Binomial, C-Logit, Proportional, Multinomial Logit, and Fixed. The first four algorithms use a predefined cost function and sample a route with the corresponding random distribution. The last algorithm only considers the travel time in free-flow condition and makes greedy choices, in which most vehicles use the route with shortest travel time for each OD.

The first dataset consists of data with three different demand patterns. The first demand pattern is called "\textbf{Single-OD}" pattern. The Single-OD pattern has only one origin source and one destination sink as shown in Figure \ref{fig:aimsun_single}. The origin source is connected to the Link 252, and the destination sink is connected to the Link 442. There are six possible shortest path candidates.

\begin{figure}
    \centering
    \begin{subfigure}{.4\textwidth}
      \centering
      \includegraphics[width=\textwidth]{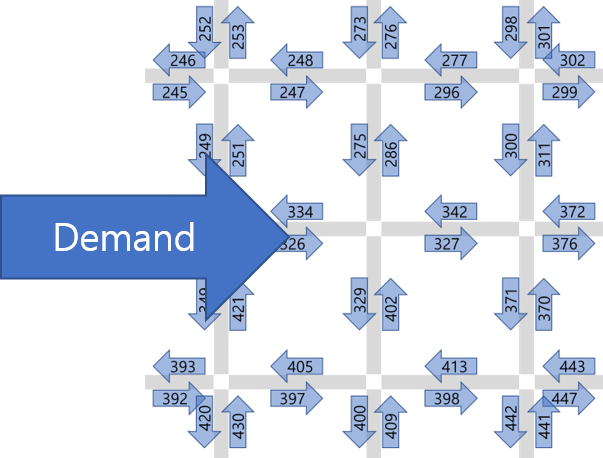}  
      \caption{One-way Multi-OD}
      \label{fig:aimsun_oneway}
    \end{subfigure}
    \hspace{6em}
    \begin{subfigure}{.45\textwidth}
      \centering
      \includegraphics[width=\textwidth]{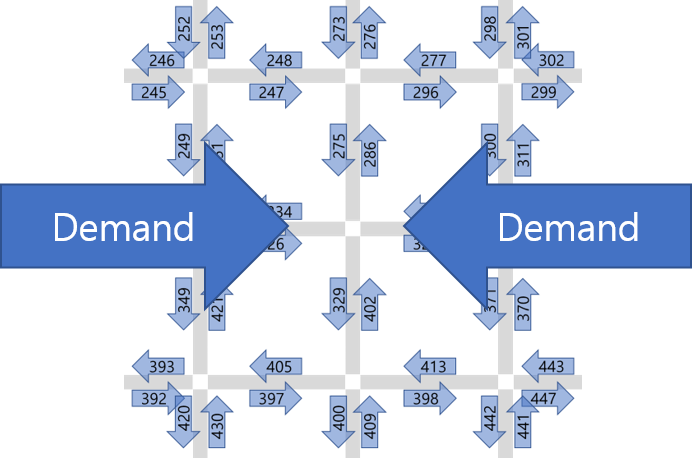}  
      \caption{Two-way Multi-OD}
      \label{fig:aimsun_twoway}
    \end{subfigure}
    \caption{Multi-OD demand patterns. Blue arrows indicates the major demand flows.}
    \label{fig:aimsun_multi}
\end{figure}

The second and third demand patterns use multiple origins and destinations. In these cases, the vehicle sources are connected to all 12 links that are at the boundary of the network, whose directions are towards the inside of the network (Link 252, 273, 298, 302, 372, 443, 441, 409, 430, 392, 321, 245). The remaining 12 boundary links, whose directions are towards the outside of the network, (Link 253, 276, 301, 299, 376, 447, 442, 400, 420, 393, 322, 246) are connected to the vehicle sinks. There can be 132 origin-destination pairs, excluding direct U-turns from the origin such as Link 245 to Link 246. The second demand pattern is called "\textbf{One-way Multi-OD}" pattern, where we assume there are major demand flows from the origin links on the left  (Link 245, 321, 392) to the destination links on the right (Link 299, 376, 447) as shown in the Figure \ref{fig:aimsun_oneway}. The major flows include all combinations of the origins (left) and destinations (right). The third demand pattern is called "\textbf{Two-way Multi-OD}" pattern, shown in Figure \ref{fig:aimsun_twoway}, where major demand flows from the origin links on the right (Link 302, 372, 443) to the destination inks on the left (Link 246, 322, 393) are added to the One-way Multi-OD pattern.

\begin{figure}
  \centering
  \includegraphics[width=0.3\textwidth]{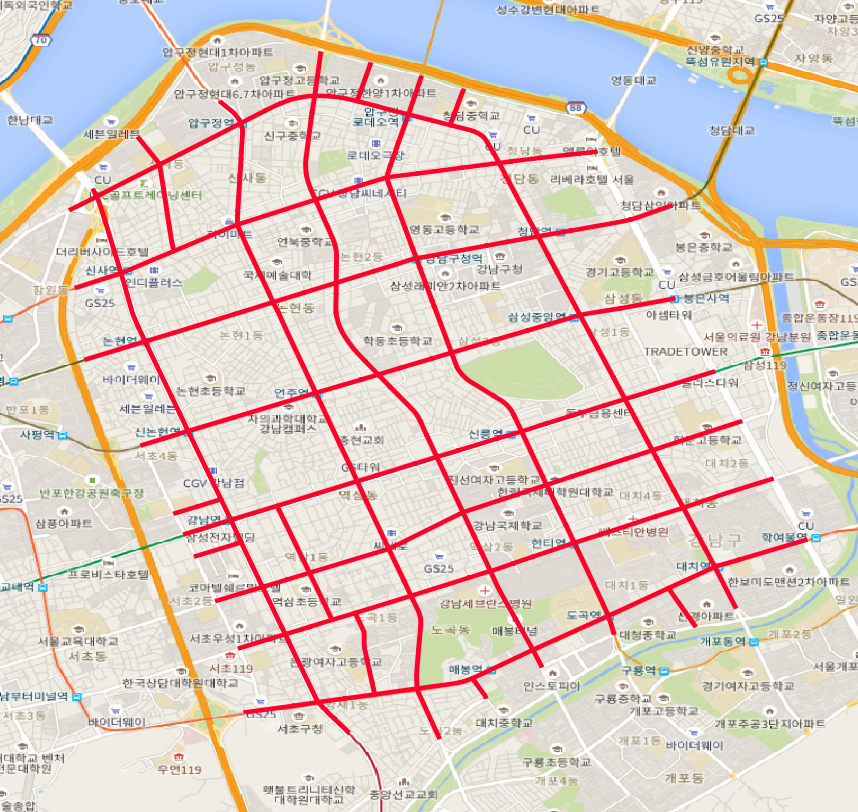}
      \caption{Major road links in Gangnam district (Map data ©2020 SK Telecom)}
  \label{fig:gangnam}
\end{figure}

The second dataset is collected by the DTG installed in taxis. The taxis operating in Seoul city are installed with digital tachographs (DTG) and collect the driving records such as position (longitude and latitude), speed, and passenger occupancy. By linking the data points with the same taxi ID in chronological order, the raw data points are transformed into a taxi trajectory dataset. Among the Seoul taxi trajectories, the trajectories that passed Gangnam district are selected for this study. Gangnam district has major road links in a grid structure as shown in Figure \ref{fig:gangnam}, so there are multiple choices in routes with similar travel distance for a given OD within the district.
%
%
Since each taxi trajectory includes multiple trips associated different passengers, we extract a sub-trajectory covering the trip of each passenger and these passenger-level sub-trajectories are referred to as "\textbf{Gangnam}" dataset. The origin and destination of the vehicle trajectory in the Gangnam dataset represent the passenger OD demand in Gangnam district, and the routing patterns represent the route choice behaviors of the taxi drivers. As mentioned in Section \ref{ProbForm}, trajectory data need to be converted into link sequences. We use a map matching algorithm in \cite{brakatsoulas2005map} to map taxi trajectory data points to the underlying road links. After the data preparation, we obtained a total of 59,553 vehicle trajectories collected in Gangnam district on January 1st, 2018.


\subsection{Baseline Models}\label{sec:baseline}

We tested TrajGAIL against three baseline models:

\begin{itemize}

\item Mobility Markov Chain

Mobility Markov Chain (MMC) \citep{gambs2010show} is one of the earliest models for the next location prediction problem based on the Markov model. MMC models each vehicle's trajectory as a discrete stochastic process, where the probability of moving to a next location depends on the immediate previous link observation. 

\begin{align}
    & P(o_{next}=i|o_{prev}=j) = \frac{N(o_{next} = i |o_{prev}=j)}{\sum_{\forall k \in O } N(o_{next} = k|o_{prev}=j)} &
\end{align}

\noindent
where $N(o_{next}=i|o_{prev}=j)$ is the number of occurrences in the real vehicle trajectory dataset that a vehicle moves from link $j$ to link $i$

\item{ Recurrent Neural Network model for Next Location Prediction}

Several previous studies such as \cite{liu2016predicting, choi2018network, choi2019attention} suggest that recurrent neural networks show good performance in predicting the next location by learning spatio-temporal features of trajectory data. When a vehicle trajectory is given, RNN cells repeatedly process and calculate the hidden state. The RNN cells decide which information to keep and which to forget. The RNN cell then calculates the probability of the next location. The cross-entropy loss is used to calculate the estimation error. When the RNN model is used as a generative model, an input vector, starting with the virtual token $Start$, is passed through RNN to compute the predictive probability over the possible next locations. One location is sampled with multinomial distribution, and the sampled next location is used as the next input vector. The procedure continues until the current location reaches the virtual token $End$ representing the end of the trip. In this study, we use Long Short Term Memory (LSTM) \citep{hochreiter1997long} for RNN cells.



\item{Maximum Entropy Inverse Reinforcement Learning} 

Maximum Entropy IRL (MaxEnt) \citep{ziebart2008maximum} is one of the most widely used IRL models. MaxEnt uses a probabilistic approach based on the principle of maximum entropy to resolve the ambiguity in choosing distributions over decisions. MaxEnt uses a linear reward function for simplicity and uses a training strategy of matching feature expectations between the observed expert policy and the learner's behavior. In \cite{ziebart2008maximum}, the feature expectation is expressed in terms of expected \textit{state visitation frequency}, meaning that the MaxEnt model calculates the expected number of visitation at each state (link in a road network in this study), and matches it with the actual number of visitation in the expert dataset. In this study, we extend the idea of matching state visitation to matching state-action visitations. We call the original MaxEnt model using state visitation frequency \textit{MaxEnt(SVF)} and the new one using state-action visitation frequency \textit{MaxEnt(SAVF)}.

\end{itemize}


\subsection{Results}\label{sec:result}

This section shows the training and testing results of the TrajGAIL and the baseline models. 
TrajGAIL and the baseline models are trained for AIMSUN and Gangnam datasets. And they  are tested in various aspects with different performance measures. Two different evaluation levels are defined to measure trajectory-level similarity and dataset-level similarity.
Section \ref{sec:training} shows the training result using convergence curves. Section \ref{sec:tra_eval} and Section \ref{sec:data_eval} show the testing result based on the trajectory-level similarity and dataset-level similarity.

\subsubsection{Training Procedure} \label{sec:training}
For the AIMSUN-based datasets, we first generate 20,000 vehicle trajectories for each demand scenario. Then, we split the total dataset into \textit{training} and \textit{testing} datasets in 0.7:0.3 ratio. As a result, we obtain 14,000 vehicle trajectories for training and 6,000 vehicle trajectories for testing. For the Gangnam dataset, we used the same ratio of 0.7:0.3, and this makes 41,687 vehicle trajectories for training and 9,866 vehicle trajectories for testing. The training dataset is only used to train each model, and all the results in Section \ref{sec:tra_eval} and Section \ref{sec:data_eval} are based on the testing dataset.

As mentioned in \textbf{Training Techniques} in Section \ref{sec:trajgail}, it is important to use sufficient number of sample trajectories at each training iterations. We tested different numbers of sample trajectories for each dataset and concluded that 2,000 sample trajectories are enough for the Single-OD dataset and 20,000 sample trajectories are enough for the Multi-OD datasets and Gangnam datasets. For a proper comparison of the model's performance, all models generate 20,000 sample trajectories for both trajectory-level evaluation and dataset-level evaluation. 
The more details on the hyperparameters used for training TrajGAIL is shown in Table \ref{tab:hypparam}.

\begin{table}
\caption{Hyperparameters used for TrajGAIL}
\label{tab:hypparam}
\begin{center}
\begin{tabular}{l l}
\hline
Hyperparameter & Value  \\ \hline
Number of iterations & 20,000      \\
Number of samples & 20,000      \\
Number of discriminator updates & 2        \\
Number of generator updates & 6        \\
Number of hidden neurons in each layer & 64 \\
Number of layers in RNN embedding & 3 \\
Learning rate & 0.00005 \\
Discount rate of reward ($\gamma$ in Eq. (\ref{eq:grad_value})) & 0.95 \\
Entropy coefficient ($\lambda$ in Eq. (\ref{eq:grad_policy}))   & 0.01 \\
\hline
\end{tabular}
\end{center}
\end{table}

\begin{figure}[ht!]
  \centering
  \includegraphics[width=0.7\textwidth]{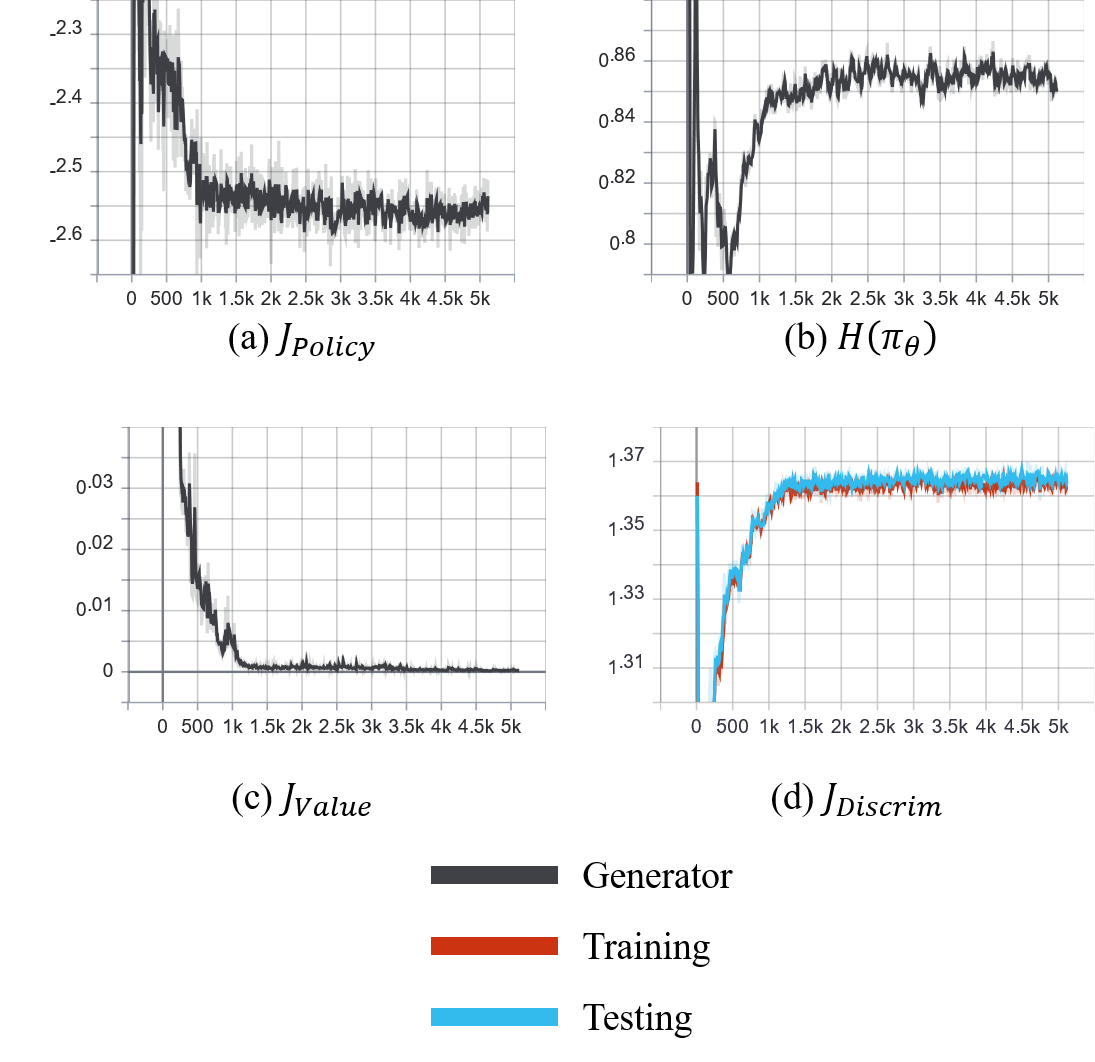}
  \caption{ Convergence curve of objective functions based on a sample case of "One-way Multi OD Binomial" using the AIMSUN dataset.}
  \label{fig:traintest}
\end{figure}

Figure \ref{fig:traintest} shows the convergence curve of the objective functions ($J_{Policy}, J_{Value}, J_{Discrim}$) and the causal entropy ($H(\pi_\theta )$) for the "One-way MultiOD Binomial" case based on the AIMSUN dataset. At the very beginning of the iterations (0 to 100 iterations), $J_{Policy}$ increases and $J_{Discrim}$ decreases as the policy generator is designed to maximize $J_{Policy}$ and the discriminator is designed to minimize $J_{Discrim}$. As the policy generator starts to produce more realistic trajectories, it gets difficult to discriminate from the perspective of the discriminator. As the discriminator starts to distinguish the real trajectories from the generated trajectories, it gets difficult for the generator to generate more realistic trajectories. As a result, $J_{Policy}$ tends to decrease and $J_{Discrim}$ tends to increase in the middle of the iterations (100 to 1000 iterations). Afterwards, $J_{Value}$ is almost converged to zero, and the generator and the discriminator makes small changes to win the minimax game. The entropy $H(\pi_\theta )$ is converged to maximize the causal entropy at this point. It is noticeable that $J_{Discrim}$ is converged to a value close to 1.38. According to \cite{goodfellow2014generative}, the discriminator objective ($J_{Discrim}$) converges to $log4 = 1.38629\cdots$ as the generator produces realistic outputs. This is because the discriminator cannot distinguish the real trajectories from the generated trajectories and, thus, the best strategy becomes a random guess, which gives 50:50 chance of getting it right.


\begin{figure}[ht!]
  \centering
  \includegraphics[width=0.5\textwidth]{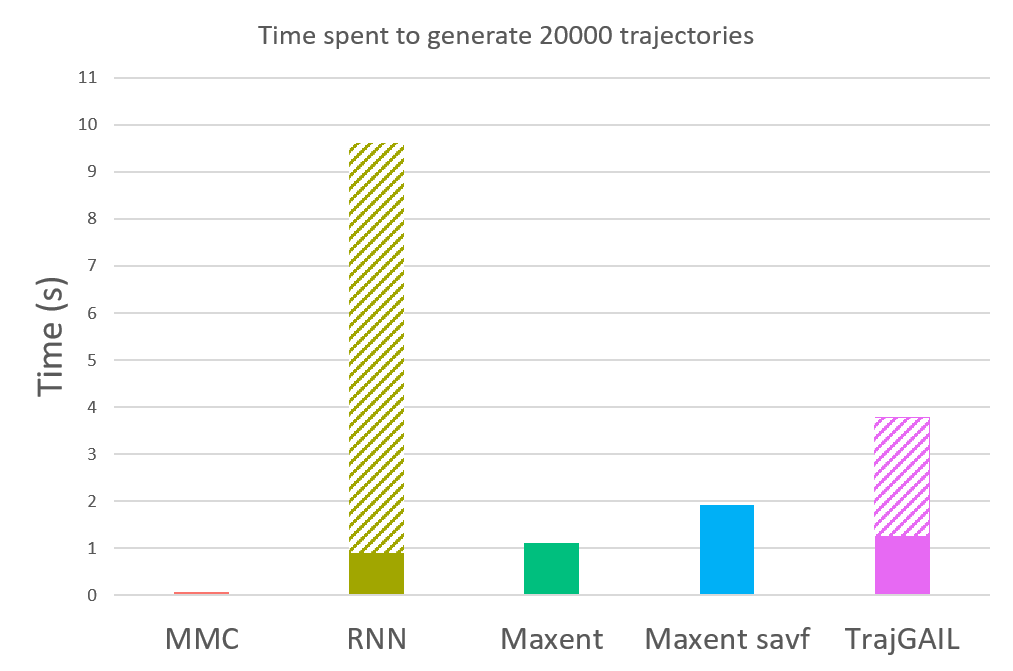}
  \caption{Computation time taken to generate 20,000 vehicle trajectories.}
  \label{fig:comptime}
\end{figure}

Figure \ref{fig:comptime} shows the computation time taken to generate 20,000 vehicle trajectories using five different models. This computation time is measured on a workstation with i9-10900KF CPU, 64GB RAM, and Nvidia Geforce RTX 3080. MMC and MaxEnt models only used CPU when computation time is measured. The computation times of RNN and TrajGAIL are measured not only by using CPU, but also by using GPU. In Figure \ref{fig:comptime}, the shaded parts of the bar chart are the computation time gain by using parallel processing of GPU when using RNN and TrajGAIL. Overall, all five models have the capability to generate 20,000 vehicle trajectories in less than 2 seconds.


\subsubsection{Trajectory-level Evaluation}\label{sec:tra_eval}

In the trajectory-level evaluation, we measure how similar each generated vehicle trajectory is to a real trajectory. Two widely used evaluation metrics in sequence modeling are used to evaluate this trajectory-level similarity: BLEU score \citep{papineni_bleu:_2002}  and METEOR score \citep{banerjee_meteor:_2005}.

In the next location prediction problem, it is common to use the probability of correctly predicting the next location to measure the model's performance.  For example, in \cite{choi2018network}, a complementary cumulative distribution function of the prediction probability is used to measure how accurately the model predicts the next one, two, or three consecutive locations. While this measure is intuitive and easy to interpret, it has  a drawback that it only considers element-wise prediction accuracy and does not take the whole sequence into account. The element-wise performance measures can be sensitive to small local mispredictions and tend to underestimate the model's performance.  As such, this study employs a BLEU score and METEOR score that consider the whole sequence. They are more robust and accurate as a performance measure for sequence modeling. 

BLEU is one of the most widely used metrics in natural language processing and sequence-to-sequence modeling. When reference sequences are given, BLEU scans through the sequence and checks if the generated sequence contains identical chunks, or a contiguous sequence of $n$ elements found in the reference sequences. Here, BLEU uses a modified form of precision to compare a reference sequence and a candidate generated sequence by clipping. For the generated sequence, the number of each chunk is clipped to a maximum count ($m_{max}$) to avoid generating the same chunks to get a higher score.

\begin{align}
\label{eq:bleu_precision}
& P_n=\frac{ \sum_{i \in C}   \min{(m_i,m_{i,max})}   }{w_t} &
\end{align}

\noindent
where $n$ is the number of elements considered as a chunk; $C$ is a set of unique chunks in the generated sequence; $m_i$ is the number of occurrences of chunk $i$ in the generated sequence; $m_{i,max}$ is the maximum number of occurrences of chunk $i$ in one reference sequence; and $w_t$ is the total number of chunks in the generated sequence.

${BLEU}_{n}$ score is defined as a multiplication of the geometric mean of $P_n$ and a brevity penalty. A brevity penalty is used to prevent very short candidates from receiving too high scores. 

\begin{align}
\label{eq:bleu}
& {BLEU}_{n} = min\Big(1, \frac{L_{gen}}{L_{ref,close}}  \Big) \cdot \Big(  \prod_{i=1}^{n} P_i  \Big)^{\frac{1}{n}} & 
\end{align}

\noindent
where $L_{gen}$ represents the length of the generated sequence, and  $L_{ref,close}$ represents the length of a reference sequence that has the closest length to the generated sequence.

METEOR \citep{banerjee_meteor:_2005} is originally designed as an evaluation metric for machine translation. It can measure similarities in terms of both the occurrences of trajectory elements and the alignment of the elements in a trajectory. METEOR first creates an alignment matching between the generated sequence and the reference sequence. The alignment matching is a set of mappings between the most similar sequence elements. Since it is often used for natural language processing, the most similar sequence element refers to the exact match, synonyms, and the stems of words. In this study, it is difficult to define the "similar" observation and state, so we only use the exact match in the alignment matching. In alignment matching, every element in the candidate sequence should be mapped to zero or one element in the reference sequence. METEOR chooses an alignment with the most mappings and the fewest crosses (fewer intersections between mappings). Based on the chosen alignment, a penalty term is calculated as follows:

\begin{align}
\label{eq:meteor_penalty}
    & p=0.5 \Big( \frac{c}{w_{map}} \Big)^3 &
\end{align}

\noindent
where $c$ is the number of chunks of elements with no crossings, and $w_{map}$ is the number of elements that have been mapped. Then, we calculate the weighted harmonic mean between precision $P$ and recall $R$ with a ratio of the weights, 1:9.

\begin{align}
\label{eq:meteor_mean}
    & F_{mean}  = \frac{10}{\frac{1}{P} + \frac{9}{R}} = \frac{10PR}{R+9P} &
\end{align}

\noindent
where $P=\frac{m}{w_{gen}}$ and $R=\frac{m}{w_{ref}}$; $m$ is the number of elements in the generated sequence that is also found in the reference sequence; and $w_{gen}$ and $w_{ref}$ are the number of elements in the generated and reference sequence, respectively.

Finally, the METEOR score, $M$, is defined as follows:

\begin{align}
\label{eq:meteor}
    & M=F_{mean}(1-p) &
\end{align}

\begin{figure}[!ht]
    \centering
    \begin{subfigure}{\textwidth}
      \centering
        \includegraphics[height=0.43\textheight]{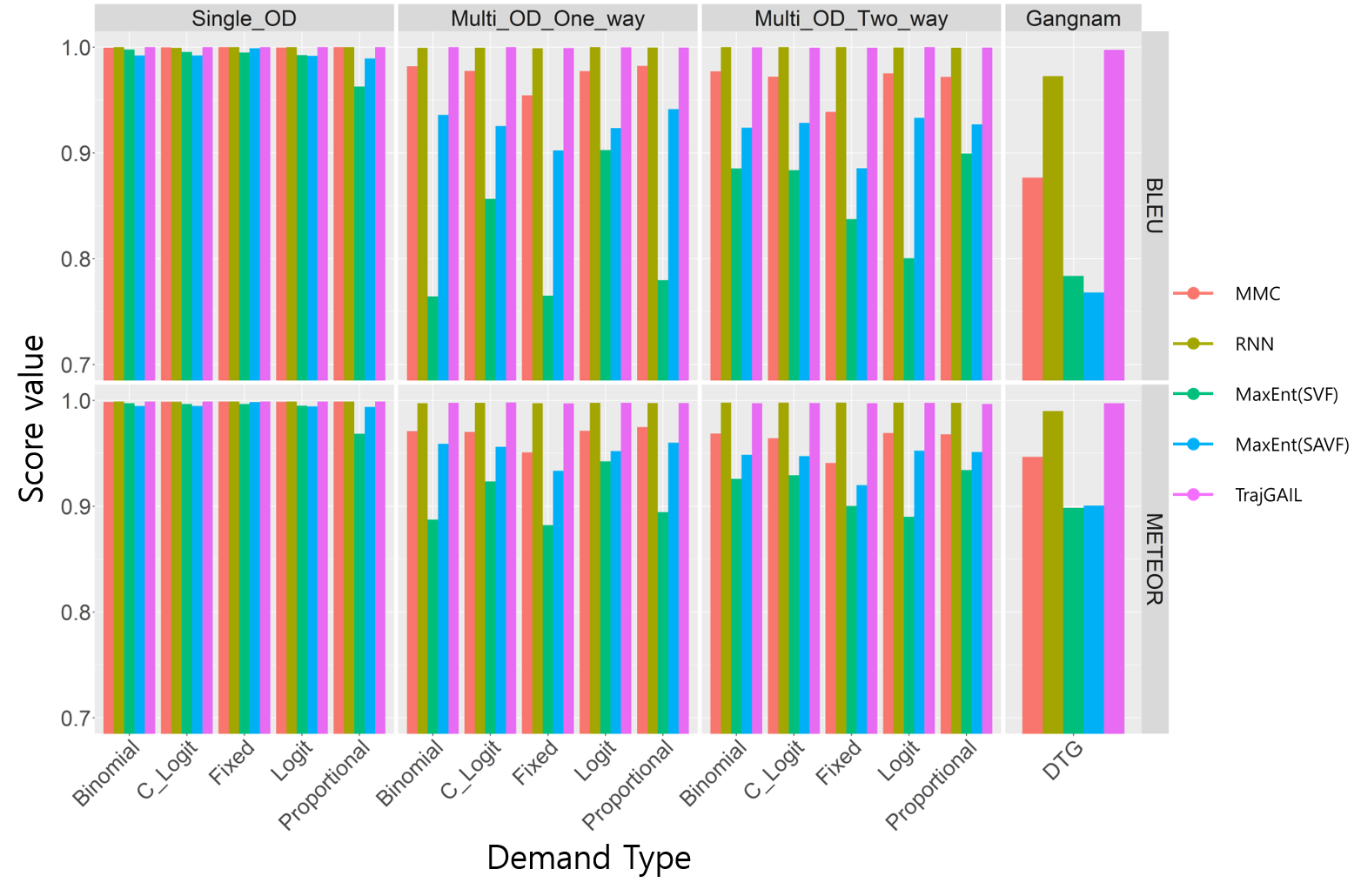}
        \caption{Average}
      \label{fig:avgscore}
    \end{subfigure}
    \par\bigskip 
    \begin{subfigure}{\textwidth}
      \centering
      \includegraphics[height=0.43\textheight]{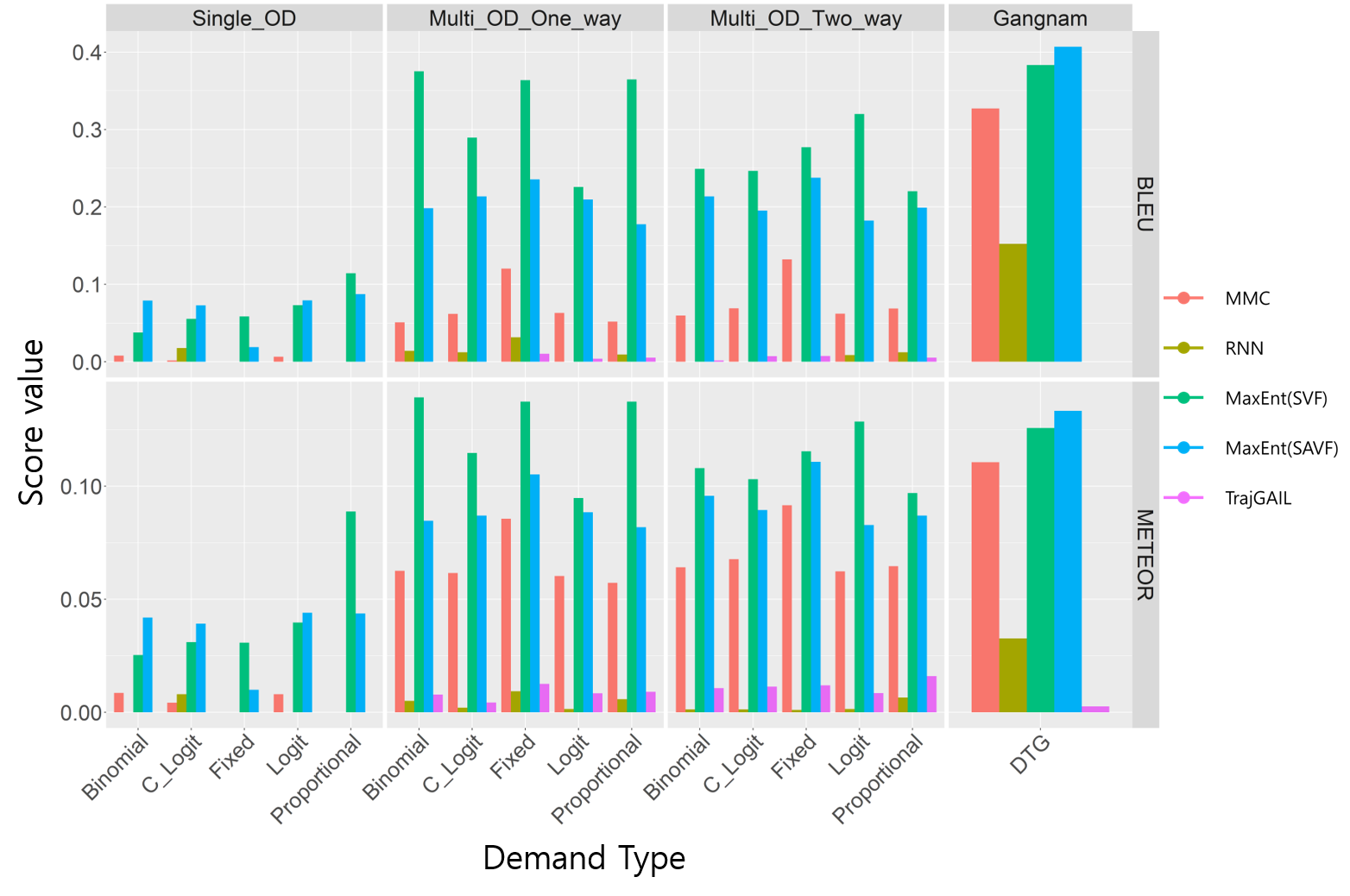}
    \caption{Standard deviation}
    \label{fig:sdscore}
    \end{subfigure}
    \caption{Result of BLEU and METEOR score of the generated vehicle trajectories of each model.}
    \label{fig:seqscore}
\end{figure}

For both BLEU and METEOR, the higher the score, the better the model performance. For BLEU score, we use $n=4$ for Eq. (\ref{eq:bleu}).
For each model, 20,000 synthetic trajectories are generated for score evaluation. 
Figure \ref{fig:seqscore} shows the result of each model in different datasets and demand types. Figure \ref{fig:avgscore} shows the average score and Figure \ref{fig:sdscore} shows the standard deviation of the score result.

When the models are tested with the Single-OD datasets, the result shows that all five models show good performance in most cases. MMC, RNN, MaxEnt(SAVF), and TrajGAIL scored more than 0.99 in both BLEU and METEOR with all five different demand types. MaxEnt(SVF) showed the lowest score (0.9627) in "Proportional" demand.


As the complexity of the dataset increases, the model's performance decreases. In Figure \ref{fig:seqscore}, MMC, MaxEnt(SVF), and MaxEnt(SAVF) show decreases in both scores when tested with the One-way Multi-OD and Two-way Multi-OD datasets. However, the scores of RNN and TrajGAIL (in the second and fifth bars within each test group) only slightly decrease. In fact, both models received perfect scores (i.e., BLEU = 1.0, METEOR = 1.0) except for few cases. In addition, when the models are tested with the Gangnam dataset, the average score of RNN and TrajGAIL is 0.9726 and 0.9974 for BLEU, respectively, and 0.9899 and 0.9974 for METEOR, respectively. Overall, the standard deviations of RNN and TrajGAIL are significantly lower than that of the other models, suggesting a higher level of robustness and lower fluctuations in model performance across different trajectory cases. The main reason for the good performance of RNN and TrajGAIL is that both models are capable of capturing sequential information in trajectories considering the history of multiple previously visited locations, as opposed to determining the next locations based only on the current location as in the other three models. However, it is observed that RNN sometimes generates trajectories traversing unknown routes (link sequences not found in the given dataset) and the relatively lower scores of RNN compared to those of TrajGAIL are attributed to instances of such unrealistic trajectories.

\subsubsection{Dataset-level Evaluation}\label{sec:data_eval}
In the dataset-level evaluation, the statistical similarity between a generated trajectory dataset and a real trajectory dataset is assessed. There are many aspects of a dataset that can be considered for statistical similarity, such as the distributions of trajectory length, origin, destination, origin-destination pair, and route. Among these variables, route distribution is the most difficult to match since producing the similar route distribution requires matching all other variables, including the lengths, origins, and destinations of vehicle trajectories in a real dataset. As such, we use a measure of route distribution similarity to evaluate dataset-level model performance.

As with the trajectory-level evaluation, each model generates 20,000 vehicle trajectories to make a synthetic trajectory dataset. We first identified all the unique routes observed in the real dataset and counted their occurrences in the synthetic dataset. 
The synthetic trajectories that travel unknown routes (i.e., the routes that did not occur in the real dataset) were marked as "unknown" trajectories. The route frequencies are calculated by dividing the route counts by the total number of trajectories in a dataset. The route frequencies, or the routes' empirical probability distribution of the synthetic dataset is compared with that of the real dataset. In this study, we use Jensen-Shannon distance $(d_{JS})$ to measure the similarity of two route probability distributions.

Jensen-Shannon distance is a widely used distance metric for two probability distributions. Given two discrete probability distributions $p$ and $q$, the Jensen-Shannon distance $(d_{JS})$ is defined as follows:
\begin{align}
 \label{eq:js}
    & d_{JS} (p,q) = \sqrt{D_{JS} (p,q)} = \sqrt{\frac{D_{KL} \Big(p||\frac{p+q}{2}\Big) + D_{KL} \Big(q||\frac{p+q}{2}\Big)}  {2}} &
 \end{align}

\noindent
where $D_{JS}$ is the Jensen-Shannon divergence, and $D_{KL}$ is the Kullback-Leibler divergence. The Kullback-Leibler divergence from $q$ to $p$, $D_{KL} (p||q)$, is defined as:

\begin{align}
  \label{eq:kl}
    & D_{KL} (p||q) = E \big[ \log (p_i) - \log (q_i) \big]  = \sum_i p_i \log \frac{p_i}{q_i} &
\end{align}

$D_{KL} (p||q)$ is also known as the relative entropy of $p$ with respect to $q$. Since $D_{KL}$ is an asymmetric similarity measure, it cannot be used as a distance metric. As a result, $d_{JS}$, a symmetrized version of $D_{KL}$, is often used to measure the distance between two probability distributions. The value of $d_{JS}$ ranges from 0 to 1, where $d_{JS} = 0$ happens when the two probability distributions are identical  and $d_{JS} = 1$ happens when the two distributions are completely different.


\begin{table}
\caption{Jensen-Shannon distance ($d_{JS}$) of Route Distribution}
\label{tab:route}
\begin{center}
\resizebox{\textwidth}{!}{
\begin{tabular}{c c c c c c c}
\hline
Dataset & Demand Type & MMC  & RNN  & MaxEnt(SVF) & MaxEnt(SAVF) & TrajGAIL    \\ \hline
SingleOD& Binomial      & 0.0866 & \textbf{0.0606}  & 0.0903 & 0.0748 & 0.0916 \\
SingleOD& C-Logit       & 0.0381          & 0.0527   & 0.1145 & 0.0650 & \textbf{0.0275} \\
SingleOD& Proportional  & \textbf{0.0192} & 0.0599   & 0.2364 & 0.0568 & 0.0274     \\
SingleOD& Logit         & 0.0448          & 0.0526   & 0.1011 & 0.0683 & \textbf{0.0284}   \\
SingleOD& Fixed         & \textbf{0.0038} & 0.0153   & 0.0594 & 0.0490 & 0.0311 \\ 
\hline
One-way MultiOD & Binomial     & 0.2822  & 0.2446    & 0.5234 & 0.3813 & \textbf{0.2125} \\
One-way MultiOD & C-Logit      & 0.3032  & 0.2501    & 0.4666 & 0.3874 & \textbf{0.1987} \\
One-way MultiOD & Proportional & 0.2988  & 0.2604    & 0.5044 & 0.3825 & \textbf{0.2059} \\
One-way MultiOD & Logit        & 0.3375  & 0.2799    & 0.4531 & 0.3893 & \textbf{0.2163} \\
One-way MultiOD & Fixed        & 0.3763  & \textbf{0.1747} & 0.5529 & 0.4629 & 0.1791 \\ 
\hline
Two-way MultiOD & Binomial     & 0.3018  & 0.3005  & 0.5011  & 0.4042  & \textbf{0.2062}\\
Two-way MultiOD & C-Logit      & 0.3328  & 0.2587  & 0.4986  & 0.4409  & \textbf{0.2072}\\
Two-way MultiOD & Proportional & 0.3430  & 0.2739  & 0.4801  & 0.4388  & \textbf{0.2090}\\
Two-way MultiOD & Logit        & 0.3375  & 0.2833  & 0.5815  & 0.4337  & \textbf{0.2021}\\
Two-way MultiOD & Fixed        & 0.3763  & 0.1783  & 0.5694  & 0.4981  & \textbf{0.1694} \\ 
\hline
Gangnam & DTG                 & 0.4701  & 0.4823  & 0.7098 & 0.5558  & \textbf{0.4230} \\ 
\hline
\end{tabular}
}

\end{center}
\end{table}

Table \ref{tab:route} shows the result of Jensen-Shannon distance ($d_{JS}$) of route distribution tested with different models. The lower the distance value, the better the model performance. The best model (with the lowest value) in each test case (row) is marked in bold. With the Single-OD datasets, all models except MaxEnt(SVF) show good results with $d_{JS}$ less than $0.1$. The $d_{JS}$ of MaxEnt(SVF) is considerably larger than that of the other models. 
It is worth noting that MMC shows good performance, especially under the Proportional and Fixed demand patterns, which is surprising considering the simplicity of the MMC model. 
Although there are some differences in the distance measures, all models except MaxEnt(SVF) were able to reproduce a synthetic trajectory dataset with high degrees of statistical similarity to the real dataset, primarily because the travel patterns in the Single-OD dataset are very simple.

The Multi-OD datasets (One-way Multi-OD and Two-way Multi-OD) have more complex trajectory patterns than the Single-OD datasets. As a result, all models show an increase in $d_{JS}$. The increase rate is significantly large in the MaxEnt models. The main reason for this is because MaxEnt models use a simple linear function to describe reward functions and the linear reward function lacks the ability to model the complex non-linear patterns of real vehicle trajectories. Both RNN and TrajGAIL have recurrent neural networks in common to use sequential information in generating synthetic trajectory. Especially, the performance of TrajGAIL is noticeably better than the other models because it not only uses sequential embedding of visited locations but also uses the reward function from the discriminator, which offers an additional guidance for a model to produce trajectories that match the real observations.



The model performance overall decreases when tested on the Gangnam dataset, i.e., the $d_{JS}$ values of all five models are above 0.4. TrajGAIL, however, still shows the best performance among the five models. The main reason for relatively large $d_{JS}$ values is that the real-world trajectory patterns in the Gangnam dataset are more complex and sparse than the simulated trajectory patterns in the other datasets. For instance, there are many rare routes with counts less than 25 among 59,553 trajectories in the Gangnam dataset. Such rare routes are difficult for a model to learn due to the limited sample sizes and, consequently, the models end up generating "unknown" trajectories when attempting to reproduce these rare routes. Among 20,000 generated trajectories, the number of unknown trajectories generated by TrajGAIL is 181, while the other four models generate more than 2,000 unknown trajectories.

\begin{figure}
\centering
\includegraphics[width=\textwidth]{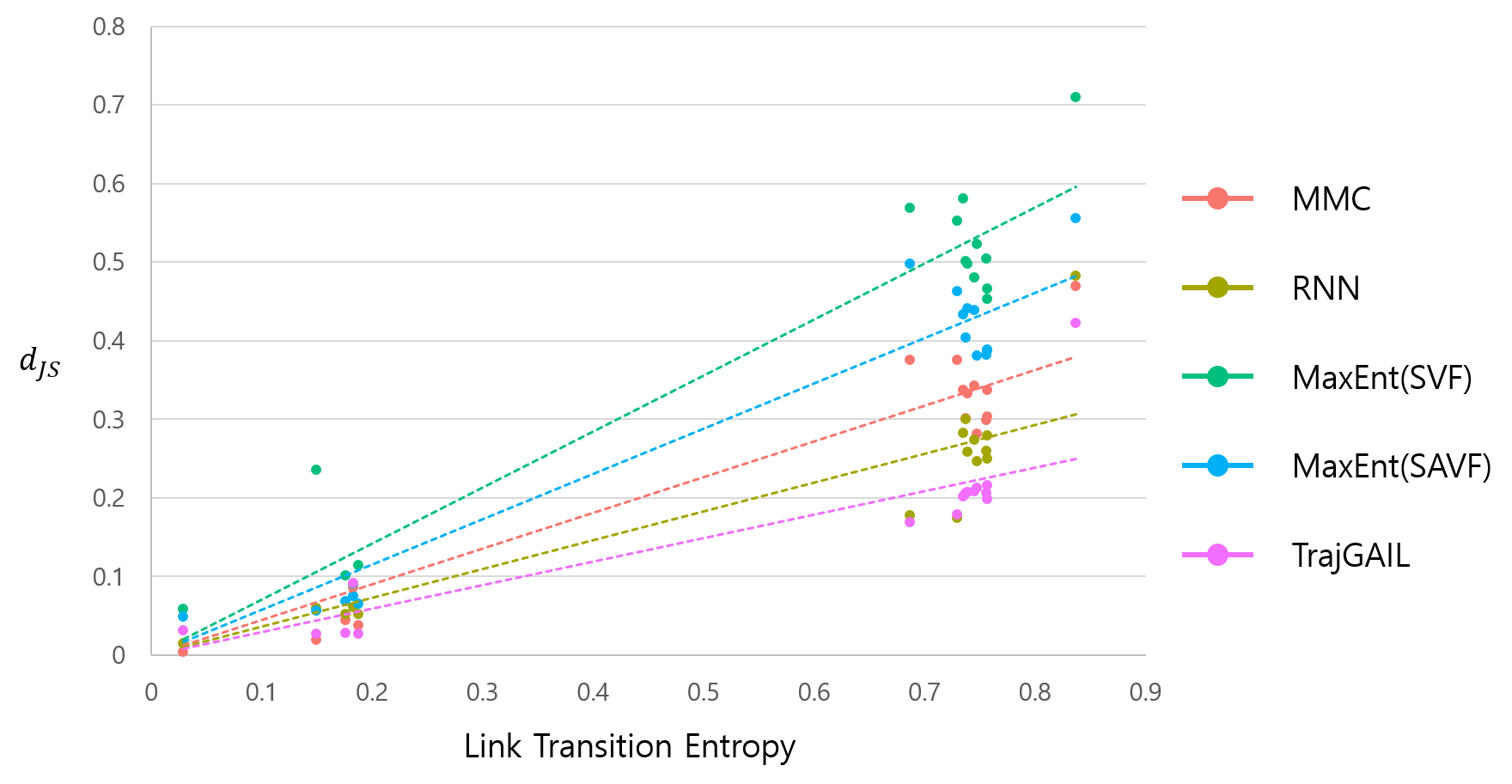}
\caption{Relationship between the link transition entropy and the Jensen-Shannon distance.}
\label{fig:entropy_jsd}
\end{figure}

In Table \ref{tab:route}, there is a tendency that $d_{JS}$ increases as the complexity of the dataset increases. To further investigate how the dataset complexity affects a model performance, we use information entropy to quantify the complexity of a dataset. More specifically, we define the \textit{link transition entropy} of a given vehicle trajectory dataset as follows:

\begin{align}
  \label{eq:entropy}
     & H(D) = \frac{1}{|L|} \sum_{l_i \in L} \Big(  \sum_{l_j \in L} -P(l_j|l_i) \log P(l_j|l_i) \Big) &
\end{align}

\noindent
where $H(D)$ is the link transition entropy of dataset $D$, $L$ is a set of possible links in $D$, and $P(l_j|l_i)$ is an empirical transition probability that a vehicle moves from link $l_i$ to link $l_j$. The intuition of this measure is to represent the dataset complexity in term of how complex or unpredictable a vehicle movement at each intersection is. For instance, a dataset would have a low $H(D)$ if all vehicles move in one direction at every intersection (e.g. all vehicles go straight at one intersection, all vehicles turn right at another intersection, and so forth), and a dataset would have a high $H(D)$ if all vehicles randomly choose the directions at all intersections.

Figure \ref{fig:entropy_jsd} shows the relationship between the link transition entropy and the Jensen-Shannon distance of route distribution from each model. In the figure, the entropy-distance results of each model are fitted into a linear model. The slope of the linear model is defined as the \textit{complexity sensitivity} of each model. A model with a high complexity sensitivity has a more significant drop in model performance as a dataset becomes more complex. On the other hand, a model with a low complexity sensitivity can learn the trajectory patterns regardless of the complexity of the dataset. The results show that TrajGAIL shows the lowest complexity sensitivity, suggesting its robustness and ability to scale to a larger and more complex datasets. RNN and MMC show the second and third lowest complexity sensitivity, while the two MaxEnt models have the highest complexity sensitivity.

\subsubsection{Discussion}

Overall, MaxEnt(SVF) showed poor performance in trajectory generation. One reason for this might be because it focuses on matching the state visitation frequency, which only cares about the element-wise similarity rather than the route-level similarity. When MaxEnt is trained to match the state-action visitation frequency, i.e., MaxEnt(SAVF), the performance improves to the level similar to the other models, which implies the importance of considering sequential information reflecting vehicles' link-to-link transition actions in generating realistic trajectories. The similar mechanism is also used in TrajGAIL, where the discriminator of TrajGAIL calculates the immediate reward based on the current state and the sampled action and this discriminator reward works in a similar way to matching the state-action visitation frequency. 

From the evaluation results, TrajGAIL and RNN are found to be the most suitable models for synthetic trajectory generation. TrajGAIL, however, outperforms RNN in most of the test cases, especially in matching route distributions to the real vehicle trajectory datasets. It is, thus, worth discussing in more detail the difference between RNN and TrajGAIL and how TrajGAIL addresses the limitations of RNN. 
Overall, the generator of TrajGAIL is similar to the RNN model. However, TrajGAIL's generator has better performance than the RNN model. 
The difference comes from the modeling architecture, i.e., how the model is designed to give proper training signals to the generator.
RNN is trained to minimize the cross-entropy loss between the probability of the predicted next location and the real next location as a label. This error is backpropagated through the input sequence to update model parameters to predict the next location based on the previously visited locations. While RNN can incorporate the previous locations into the next location prediction, it does not consider the rest of the trip. 
In contrast, TrajGAIL uses the reward function from the discriminator combined with the value estimator to consider the rest of the trip. The reward function from the discriminator captures how realistic the current state and action are. The value estimator then calculates the $\gamma$-discounted cumulative rewards, which capture how realistic the remaining states and actions will be. Using these two functions, the generator of TrajGAIL can learn a more comprehensive distribution of a given trajectory dataset and generate trajectories reflecting the sequential patterns along the whole trajectories in real data.

\section{Conclusion}\label{sec:conclusion}
This study proposes TrajGAIL, a generative adversarial imitation learning framework for urban vehicle trajectory generation. In TrajGAIL, drivers' movement decisions in an urban road network are modeled as a partially observable Markov decision process (POMDP). The generative adversarial imitation learning is then used to learn the underlying decision process demonstrated in the given trajectory dataset. This allows TrajGAIL to generate new synthetic trajectory data that are similar to real-world trajectory observations. 

In terms of contributions, this study have made several theoretical and methodological advancements in the field of trajectory analysis and synthetic trajectory generation. TrajGAIL adopted the combination of POMDP and RNN embedding to encode the historical sequence in a current state without violating the Markov assumption required for the MDP-based imitation learning framework. This overcomes the disadvantage of standard GAIL and traditional IRL (e.g., MaxEnt) that they cannot take into account the previous locations in selecting the next location in a trajectory. TrajGAIL demonstrated the superiority of GAIL over GAN in generating location sequences from a generative adversarial framework. The imitation learning approach in GAIL allows the consideration of not only the previous locations but also future locations to make the whole trajectories more realistic, while GAN does not take into account future locations and is prone to generate unrealistic trajectories with excessive lengths and several loops. In terms of technical contributions, the use of deep learning approach (i.e., GAIL) over traditional IRL in the context of imitation learning overcomes the IRL's drawback of high computation cost when dealing with a large state-action space and, thus, offers better scalability for the applications using large-scale traffic networks.


The model's performance is evaluated on different datasets with various traffic demand patterns against three baseline models from previous studies. The evaluation is divided into two levels: trajectory-level evaluation and dataset-level evaluation.
In the trajectory-level evaluation, the generated vehicle trajectories are evaluated in terms of BLEU and METEOR, the two most widely used scores in sequence modeling. In the dataset-level evaluation, the statistical similarity between the generated vehicle trajectory dataset and the real vehicle trajectory dataset is measured using the Jensen-Shannon distance of route distribution. The results show that TrajGAIL can successfully generate realistic trajectories that capture trajectory-level sequence patterns as well as match route distributions in the underlying datasets, evidenced by significantly higher performance measures of TrajGAIL compared to the other models. A model's performance sensitivity with respect to the complexity of a trajectory dataset was further investigated by measuring the link transition entropy of the dataset and analysing its relationship with model performance. The results show that TrajGAIL is least sensitive to dataset complexity among the tested models, suggesting its robustness in learning complex patterns of real vehicle movements and ability to scale to a larger and more complex dataset. 



There are several directions in which the current study can be extended to further improve the trajectory generation performance. Currently, we only encode the sequence of links into the belief states in POMDP as a way to incorporate unobserved variables in modeling and predicting next locations. There are, however, other variables that can help sequence prediction in addition to the visited link sequence information. For instance, traffic conditions in a road network, vehicles' origin and destination information, and temporal information such as time-of-day and day-of-week can all provide additional information to further improve trajectory prediction and generation. While the current paper did not consider this as we focus on introducing the theoretical aspects of TrajGAIL and analyzing the effects of the structural differences between TrajGAIL and other models, we will consider incorporating other variables in our future study.

Additional information can be incorporated into TrajGAIL in various ways. \cite{choi2019attention} used an attention-based RNN model to incorporate network traffic states into the next location prediction problem. A similar attention mechanism can be employed to give the network traffic state information to the generator of TrajGAIL. As human drivers use traffic state information from navigation services to make better decision in route choices, this attention mechanism can guide the generator to pay attention to the traffic states of certain locations in the road network to select actions more accurately, in a way that human drivers do.

Another way is to use a conditional version of generative adversarial modeling framework such as conditional-GAN (cGAN) \citep{mirza2014conditional} and conditional-GAIL (cGAIL) \citep{zhang2019unveiling}, which allow models to generate synthetic data conditioned on some extra information. 
For example, we can feed trajectories' "origin" locations as additional input to TrajGAIL during training such that the generator and discriminator are trained to learn vehicle trajectory patterns given origin locations. This would enable the model to further capture and distinguish different route choice behaviors specific to different origin regions.
%
%
Instead of feeding extra information to the model in a supervised way as in cGAN and cGAIL, it is also possible to achieve this in an unsupervised way as in infoGAN \citep{chen2016infogan} and infoGAIL \citep{li2017infogail}. Instead of explicit condition inputs, InfoGAN and infoGAIL introduce latent variables, which are used by a model to automatically distinguish certain behaviors in data in a meaningful and interpretable way.
For instance, it is possible to build a trajectory generation model that can automatically cluster and distinguish patterns in trajectory data by different latent features such as origins, destinations, and time-varying traffic demands and use this information to guide the generating process.




Lastly, it is worth mentioning the connection between our TrajGAIL model and traditional route choice models in transportation such as discrete choice model and dynamic traffic assignment (DTA) because there are similarities between them in that they both aim to model an individual driver's choice behaviors in selecting routes in traffic networks. A main difference lies in how "route" and "choice" are defined. In traditional route choice models, a route is "selected" from a pre-defined set of alternative routes for a given OD and, thus, a choice is made at the OD and route level. In our generative modeling approach, on the other hand, a route is "constructed" as a result of sequential decisions (link-to-link transitions) along the journey and, thus, a choice is made at the intersection and link level. Although a decision is made locally at each intersection, the \textit{learned reward function} captures the network-wide route choice patterns and, thus, can give correct signals at each local decision point to enable the generated trajectories to exhibit realistic route- and network-level patterns globally. Another major difference lies in modeling approach. Traditional route choice models are "model-based" (or theory-based) in that they rely on behavioral assumptions and theories (e.g., utility maximization, user equilibrium) and aim to explain "why" a specific driver chooses a certain route. Our generative approach is "data-driven" and does not rely on any behavioral assumptions. It does not try to explain "why" but instead focuses "what" patterns exist in the actual realized data and "how" to reproduce them.

In sum, TrajGAIL offers a data-driven alternative to traditional route choice models for describing the underlying route distribution of a traffic network in synthetic data generation problems, without requiring the identification of ODs and the computationally expensive enumeration of route sets. There are limitations in TrajGAIL, however, that currently it does not take into account the effects of traffic conditions or interactions with other vehicles and, thus, is unable to serve as a route choice model for DTA or traffic simulation models. It is an important and interesting future research topic to consider how data-driven deep learning models could complement or replace existing theory-based modeling components to improve DTA and traffic simulation. The extensions to TrajGAIL discussed above (e.g., attention, cGAIL, and infoGAIL) can allow the consideration of additional information and, thus, could be potentially used in that direction.

\newpage



\printcredits

\bibliographystyle{cas-model2-names}

\bibliography{partc}





\end{document}